\documentclass[letterpaper]{article} 
\usepackage{aaai24}  
\usepackage{times}  
\usepackage{helvet}  
\usepackage{courier}  
\usepackage[hyphens]{url}  
\usepackage{graphicx} 
\usepackage{natbib}  
\usepackage{caption} 
\usepackage{algorithm}
\usepackage{algorithmic}

\usepackage{url}
\usepackage{amsfonts,amsmath,amssymb}
\usepackage{stix,utfsym}
\usepackage{enumitem}

\usepackage{newfloat}
\usepackage{listings}
\DeclareCaptionStyle{ruled}{labelfont=normalfont,labelsep=colon,strut=off} 
\floatstyle{ruled}
\newfloat{listing}{tb}{lst}{}
\floatname{listing}{Listing}
\usepackage{xspace,mfirstuc,tabulary}
\usepackage{comment}
\usepackage{xcolor}
\definecolor{ao(english)}{rgb}{0.0, 0.5, 0.0}
\usepackage{subcaption}
\usepackage{afterpage} 

\usepackage{booktabs}

\newcommand{\grnsq}{$\color{ao(english)}\blacksquare$}

\newcommand{\grysq}{$\color{gray}\blacksquare$}
\newcommand{\redx}{$\color{red}\textbf{X}$}

\newcommand{\gr}[1]{}    
\newcommand{\bdk}[1]{}      
\newcommand{\nilay}[1]{} 
\newcommand{\jmf}[1]{}      
\newcommand{\cm}[1]{}  
\newcommand{\jmfb}[1]{}      

\title{Task Contamination: Language Models May Not Be Few-Shot Anymore}
\author {
    Changmao Li, 
    Jeffrey Flanigan
}
\affiliations {
    University of California, Santa Cruz\\
    changmao.li@ucsc.edu, jmflanig@ucsc.edu
}

\usepackage{bibentry}

\begin{document}

\maketitle

\begin{abstract}
Large language models (LLMs) offer impressive performance in various zero-shot and few-shot tasks. However, their success in zero-shot and few-shot settings may be affected by task contamination, a potential limitation that has not been thoroughly examined. This paper investigates how zero-shot and few-shot performance of LLMs has changed chronologically over time. Utilizing GPT-3 series models and several other recent open-sourced LLMs, and controlling for dataset difficulty, we find that on datasets released before the LLM training data creation date, LLMs perform surprisingly better than on datasets released after. This strongly indicates that, for many LLMs, there exists task contamination on zero-shot and few-shot evaluation for datasets released prior to the LLMs' training data creation date. Additionally, we utilize training data inspection, task example extraction, and a membership inference attack, which reveal further evidence of task contamination. Importantly, we find that for classification tasks with no possibility of task contamination, LLMs rarely demonstrate statistically significant improvements over simple majority baselines, in both zero and few-shot settings.
\end{abstract}


\section{Introduction}

Recently there has been much interest in few-shot methods, in particular in-context learning (ICL, Brown et al. 2020) with large language models.  In-context learning has the benefit of yielding excellent performance while requiring very little data, sometimes relying on only a few examples for the task\nilay{maybe change 'few' to '< 10' to give concrete number?}.  These promising results have led to an explosion of work on in-context learning methods across a wide variety of tasks \cite{schick-schutze-2021a, schick-schutze-2021b, poesia2022, hu-etal-2022},\jmf{shorten to just the highlights, 3-4 citations} including prompt tuning methods \cite{qin2021, lester2021}, chain-of-thought methods \cite{wei2022, wang2022, wang2023, aiyappa2023}, tool-based methods \cite{Timo2023, yang2023}.

However, along with this explosion of work in ICL, many have raised concerns about data contamination \cite{brown2020, jacovi2023}, that is, prior knowledge of data or a task which is thought to be unseen by the model.\jmf{add citation to GPT paper that did data contamination analysis.}\jmf{define data contamination} Data contamination can happen in multiple ways. One common contaminant is \textbf{test data contamination}, the inclusion of test data examples and labels in the pre-training data. Another contaminant for zero or few-shot methods, which we call \textbf{task contamination}, is the inclusion of task training examples in the pre-training data, effectively making the evaluation no longer zero or few-shot.\footnote{Zero-shot evaluation is evaluation where a model has seen zero examples for the task.  Few-shot, or $N$-shot, where $N$ is a small number, is where the model has seen $N$ examples for the task. Prior work has sometimes defined zero-shot for multi-class classification as predicting \textit{classes} that have never been seen during training, but most recent work does not use this definition.}

\begin{figure}[t]
\centering
\includegraphics[scale=1]{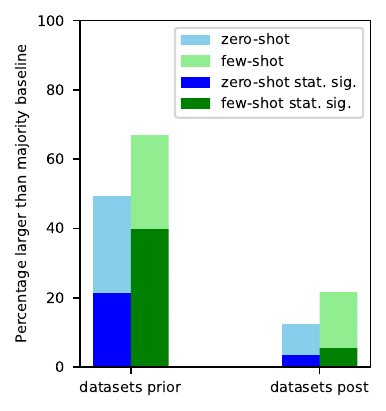}
\caption{Percentage of datasets with accuracy higher than the majority baseline for datasets released prior and post LLM training data collection date, for both zero-shot (blue, left) and few-shot (green, right).  Results are across all models and all datasets. On datasets released post training data collection date for the LLM, the LLM is much less likely to improve upon the simple majority baseline. \textit{Stat. sig.} (darker) is the percent of datasets for which the performance above majority baseline is significant at the 99\% confidence level.}
\label{fig:baseline-percentage}
\end{figure}

Simply evaluating the scope of this contamination is difficult to do \cite{ magar2022, jacovi2023}.  Closed models do not release their pre-training data.  While open models give the sources, crawling the sites to obtain that data is non-trivial, especially if the data has changed from when it was crawled.  For models that are pre-trained on freely available pre-training corpora, simply grepping for examples in the pre-training corpora may not be reliable due to differences in data formatting (such as XML vs CVS, etc) or differences in text normalization and tokenization.

In this paper we empirically measure the scope of task contamination for few-shot methods across various models and tasks. To the best of our knowledge, we are the first to systematically analyze this problem. We evaluate 12 different models, ranging from closed GPT-3 series models \cite{openai2023a} to open models including Fairseq MoE \cite{artetxe2022}, GPT-J \cite{gpt-j}, Bloom \cite{scao2022}, OPT \cite{zhang2022} ,  LLaMA \cite{touvron2023}, Alpaca \cite{alpaca}, and Vicuna \cite{vicuna2023} on 16 classification tasks and 1 semantic parsing task\nilay{Reviewers might be concerned with lack of seq2seq tasks.}.

We analyze each model on datasets created before its training data was crawled on the internet versus datasets created afterward. We find that datasets created before the LLM training data was collected have a significantly higher chance of having performance higher than the majority baseline (Fig.~\ref{fig:baseline-percentage}).

We perform training data inspection and task example extraction to look for possible task contamination.  Importantly, we find that for classification tasks with no possibility of task contamination, models rarely demonstrate statistically significant improvements over simple majority baselines across a range of tasks, in both zero and few-shot settings (Fig.~\ref{fig:experiments-majority}).


As a case study, we also attempt to conduct a membership inference attack for a semantic parsing task (Spider, Yu et al. 2019) for all models in our analysis. We find a strong correlation (R=.88) between the number of extracted examples and the accuracy of the model on the final task (Fig.~\ref{fig:dev-acc-em}).  This is strong evidence that the performance increase in zero-shot performance on this task is due to task contamination.

Additionally, we look closely at the GPT-3 series models. We find that training examples can be extracted from the GPT-3 models, and that the number of extractable training examples increased from each version from \texttt{davinci} to \texttt{GPT-3.5-turbo}, and closely tracks the increase in zero-shot performance of the  GPT-3 models on that task (Fig.~\ref{fig:experiments-majority}).  This is strong evidence that the increase in performance on these tasks across GPT-3 models from  \texttt{davinci} to \texttt{GPT-3.5-turbo} is due to task contamination.

\section{Overview}

We employ four methods of measuring task contamination.\jmf{Add a figure containing examples of each of these.}

\begin{enumerate}
\item \textbf{Training data inspection}: Search through the training data to find task training examples.

\item \textbf{Task example extraction}: Extract task examples from an existing model. Extraction is only possible with instruction-tuned models. This analysis can also be done for training data or testing data extraction~\cite{lm-contamination}. Note: For the purposes of detecting task contamination, the extracted task examples need not exactly match existing training data examples. Any examples demonstrating the task indicate possible contamination for zero and few-shot learning.

\item \textbf{Membership inference}: This method only applies to generation tasks.  Check if the model generated content for an input instance is exactly the same as the original dataset \cite{hu2022}\jmf{add inter-alia}. If there is an exact match, we can infer it is a member of the LLM's training data. This differs from task example extraction because generated output is checked for an exact match.  Exact matches for an open-ended generation task strongly indicate the model has seen those examples during training.  The model is not just good, it is psychic: it has knowledge of the exact phrasing used in the data. Note: this can only be used for generation tasks.\footnote{Exact matches for the input do not indicate task contamination because the input text could have been seen, but it needs to be paired with the output label for task contamination.}  

\item \textbf{Chronological analysis}: For a set of models whose training data has been collected at a range of known times, measure performance on a dataset with a known release date, and check for evidence of contamination using chronological evidence.

\end{enumerate}

The first three methods have high precision, but suffer from low recall.  If data is found in the training data for the task, then it is certain that it has seen examples.  But because of data formatting variations, variations in keywords used to define the task, and the size of the dataset, the absence of evidence for contamination using the first three methods is not evidence of absence.

The fourth method, chronological analysis, is high recall, but low precision.  If the performance is high due to task contamination, then a chronological analysis will have a high chance of catching it.  But other factors could also contribute to increased performance over time, so the precision is low.

Due to their inherent trade-offs, we employ all four methods for detecting task contamination.  With all four methods, we find strong evidence of task contamination for some combinations of models and datasets.  We begin with a chronological analysis for all models and datasets we tested, since it has the highest potential for catching possible contamination (\S\ref{sec:chronological}).  We then look for further evidence of task contamination using training data inspection (\S\ref{sec:trainingdatainspection}) and task example extraction (\S\ref{sec:trainingdataextraction}). Next we look at the performance of LLMs on tasks without contamination (\S\ref{sec:no_contamination}), and conclude with additional analysis using a membership inference attack (\S\ref{sec:membershipinference}).

\section{Models and Datasets}

\paragraph{Models}
\jmf{todo}
We experimented with 12 models. Table~\ref{tab:data-time} lists these models, along with the collection dates of the training data and release dates for each model.\footnote{GPT-3 series training data collection dates are obtained from \url{https://platform.openai.com/docs/models/overview}} The 12 models we use can be further categorized into two broad groups: (1) five proprietary GPT-3 series models ("closed") and (2) seven open models with free access to their weights ("open"). Comparing models from these two groups yields valuable insights into the difference between proprietary, high-performance models like those from the GPT-3 series and more accessible, community-driven open models. More information about hyperparameters for these models is given in the Appendix \ref{app:hyperparameters}.


\begin{table}[t!]
\centering 
\begin{subtable}{\columnwidth}
\centering 
\begin{tabular}{lll}
\toprule
Model            & Training data & Release   \\
\midrule
davinci          & Up to Oct 2019 & May 2020       \\
davinci-001 & Up to Oct 2019 & Jun 2020       \\
davinci-002 & Up to Jun 2021 & Jan 2022      \\
davinci-003 & Up to Jun 2021 & Nov 2022        \\
GPT-3.5-T    & Up to Sep 2021  & Mar 2023    \\
\bottomrule
\end{tabular}
\caption{GPT-3 Series LLMs}
\end{subtable}

\hspace{0.1cm}

\begin{subtable}{\columnwidth}
\centering 
\begin{tabular}{lll}
\toprule
Model            & Training data & Release   \\
\midrule
Fairseq MoE      & Up to Feb 2019 &  Dec 2021     \\
GPT-J            & Up to 2020 & Jun 2021           \\
OPT              & Up to Oct 2021 &  May 2022     \\
BLOOM            & Prior Aug 2022 &  Nov 2022     \\
LLaMA            & Up to Aug 2022  & Feb 2023    \\
Alpaca           & From davinci-003 & Mar 2023 \\
Vicuna           & From ChatGPT  & Mar 2023   \\
\bottomrule
\end{tabular}
\caption{Open LLMs}
\end{subtable}
\caption{Dates for the training data creation and model release. davinci-XXX refers to \texttt{text-davinci-XXX}. GPT-3.5-T refers to \texttt{GPT-3.5-turbo-0301}.}
\label{tab:data-time}
\end{table}

\paragraph{Datasets}
\jmf{update and expand}
Zero-shot and few-shot evaluations involve models making predictions on tasks that they have never seen or seen only a few times during training. The key premise is that the models have no prior exposure to the particular task at hand, ensuring a fair evaluation of their learning capacity. Contaminated models, however, give a false impression of its zero- or few-shot competency, as they have already been trained on task examples during pretraining. Detecting such inconsistencies would be relatively easier in a chronologically ordered dataset, where any overlap or anomaly would stand out. Based on this narrative, we split the datasets into two categories: 
datasets released before or after January 1st, 2021, identified as \textbf{pre-2021} datasets and \textbf{post-2021} datasets. We use this division to analyze the zero-shot or few-shot performance difference between older datasets and newer ones, with the same division applied for all LLMs. We also use the per-LLM division \textbf{pre-collection} and \textbf{post-collection} datasets, which distinguishes datasets that the model was possibly trained on (pre-collection datasets) from the datasets it could not have been trained on (post-collection datasets). Table~\ref{tab:data-time} presents the creation time of the training data for each model. Information about the datasets can be found in the Appendix \ref{app:datasets}, while release dates for each dataset are listed in Table~\ref{table:datasets}.

\begin{table}[htpb!]
\centering
\begin{tabular}{lllll}

\toprule
\multicolumn{2}{c}{Pre-2021} &  & \multicolumn{2}{c}{Post-2021} \\
     Dataset     &     Year     & ~~~ &     Dataset      &        Year  \\
     \cmidrule{1-2} \cmidrule{4-5}
     RTE     &    2009      &  &StrategyQA &  2021 \\
     WNLI     &   2011       &  &NewsMTSC-MT & 2021 \\
     COPA & 2011 &  & NewsMTSC-RW & 2021\\
     SST-2 & 2013&    & NLI4Wills & 2022 \\
     MRPC & 2015  &   & CREPE & 2023\\
     QNLI & 2018 &  & FOMC & 2023 \\
     CB &  2019&   &  NewsMet & 2023 \\
     WiC & 2019&   & \multicolumn{2}{c}{}\\
     BoolQ &  2019&  \\
\bottomrule
\end{tabular}
\caption{Dataset release year for each dataset, split into pre-2021 datasets and post-2021 datasets. }
\label{table:datasets}
\end{table}

\begin{figure*}[t]
        \centering
        \begin{subfigure}[b]{0.45\textwidth}
            \centering
            \includegraphics[scale=0.45]{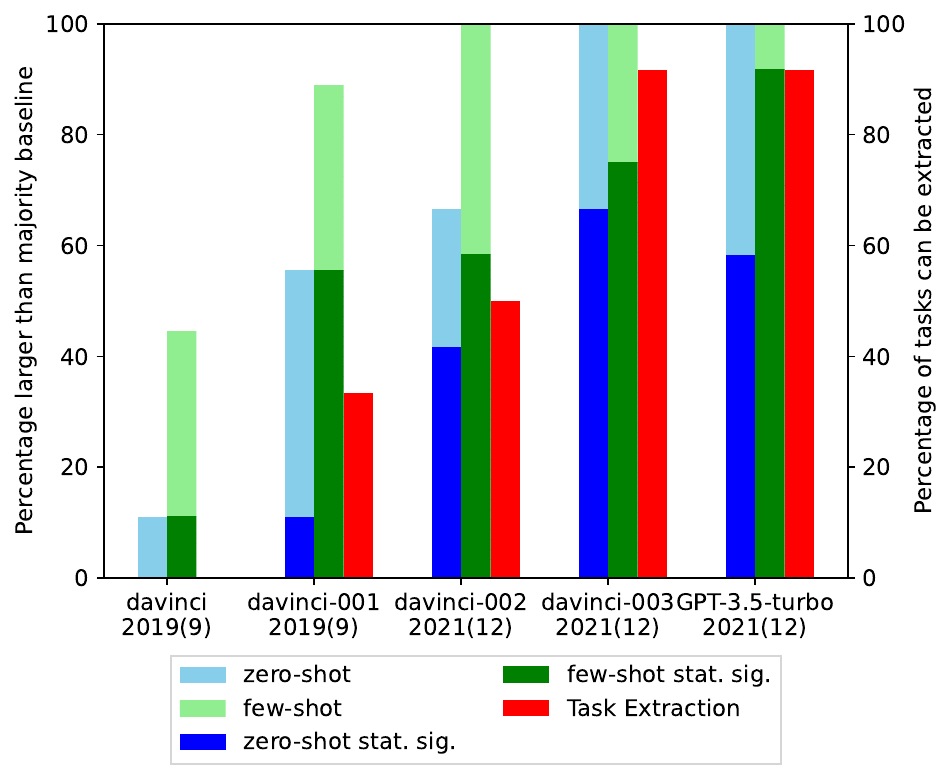}
            \caption[]%
            {GPT-3 series on pre-collection datasets}
            \label{fig:Zero shot performance for old datasets1}
        \end{subfigure}
        \begin{subfigure}[b]{0.45\textwidth}   
            \centering 
            \includegraphics[scale=0.45]{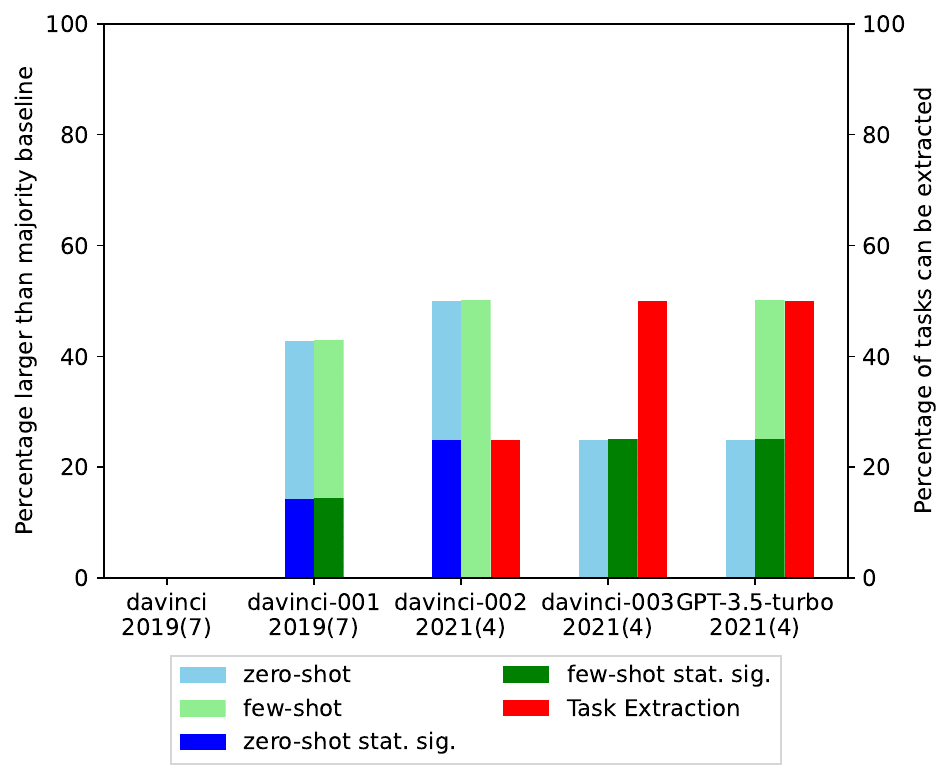}
            \caption[]%
            {GPT-3 series on post-collection datasets}    
            \label{fig:Few shot performance for old datasets1}
        \end{subfigure}
        \begin{subfigure}[b]{0.45\textwidth}
            \centering
            \includegraphics[scale=0.45]{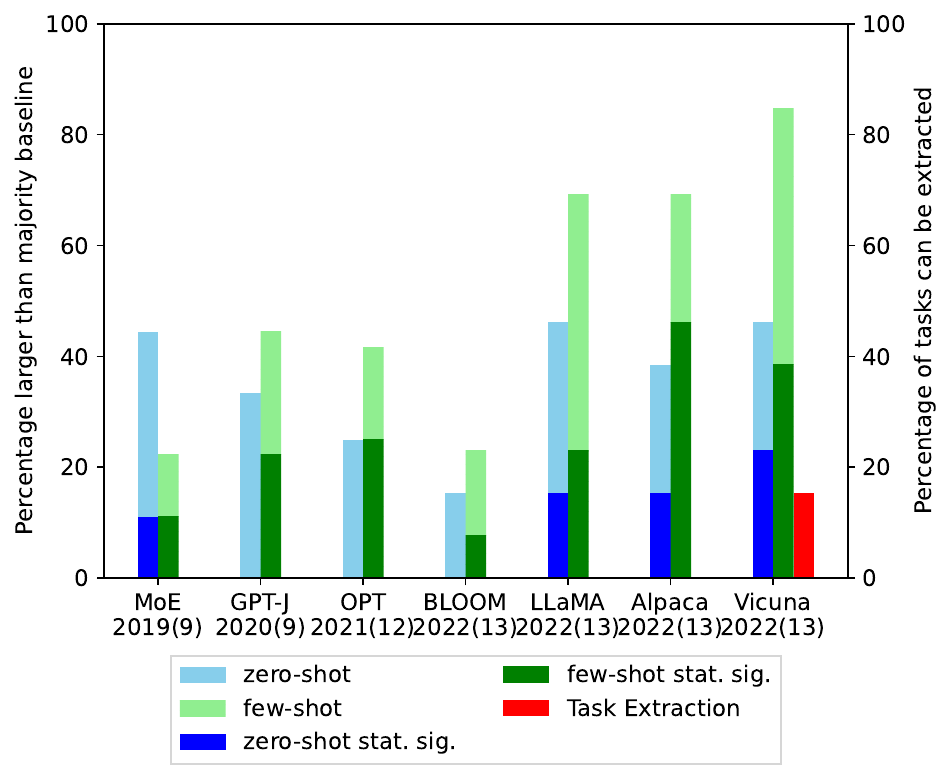}
            \caption[]%
            {Open LLMs on pre-collection datasets} 
            \label{fig:Zero shot performance for old datasets2}
        \end{subfigure}
        \begin{subfigure}[b]{0.45\textwidth}   
            \centering 
            \includegraphics[scale=0.45]{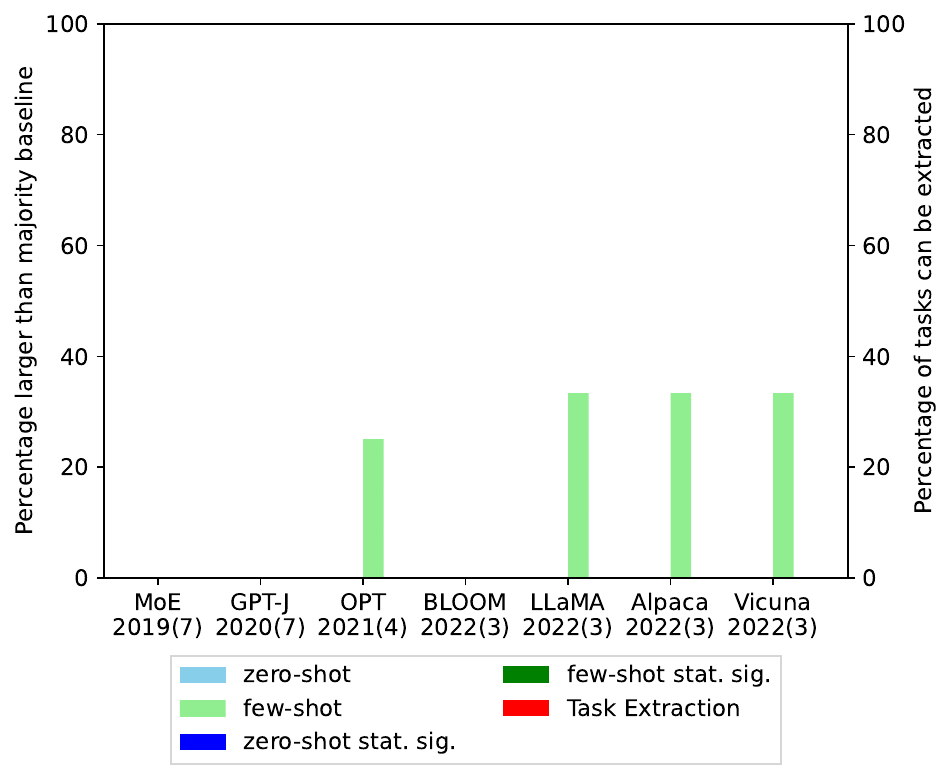}
            \caption[]%
            {Open LLMs on post-collection datasets}    
            \label{fig:Few shot performance for old datasets2}
        \end{subfigure}
        \caption[]
        {Percentage of datasets larger than majority baselines for each LLM (light color), as well as the percentage of tasks for which training data can be extracted with an instruction prompt (Red, see also Table~\ref{tab:extract}). Dark color is the percentage of datasets significantly larger ($p=.99$) than the majority baseline using a t-test. Below each LLM, we list the training data collection year, and the total number of datasets in pre- or post-collection in parenthesis (e.g. MoE has 7 datasets post training collection date.)  For tasks without demonstrated possibility of task contamination (post-collection datasets (b) and (d), with no extracted task examples in red), models rarely show statistically significant improvements over majority baselines (see \S\ref{sec:no_contamination} for details).}
        \label{fig:experiments-majority}
    \end{figure*}

\begin{figure*}[t]
        \centering
        \begin{subfigure}[b]{0.475\textwidth}
            \centering
            \includegraphics[scale=0.45]{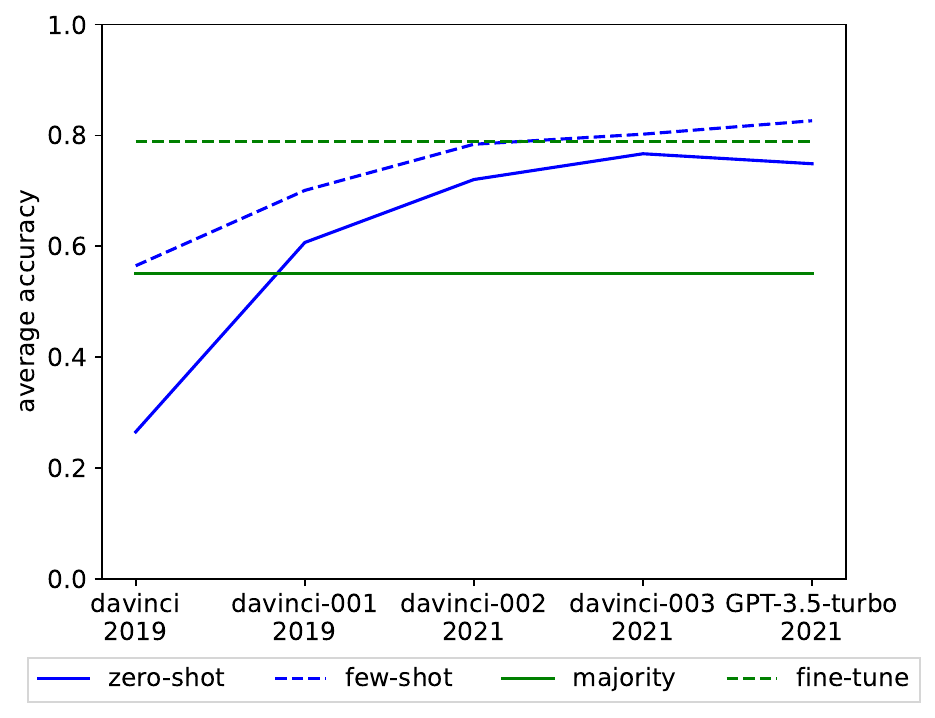}
            \caption[]%
            {{GPT-3 series on pre-2021 datasets.}}    
            \label{fig:GPT-old-datasets}
        \end{subfigure}
        \begin{subfigure}[b]{0.475\textwidth}   
            \centering 
            \includegraphics[scale=0.45]{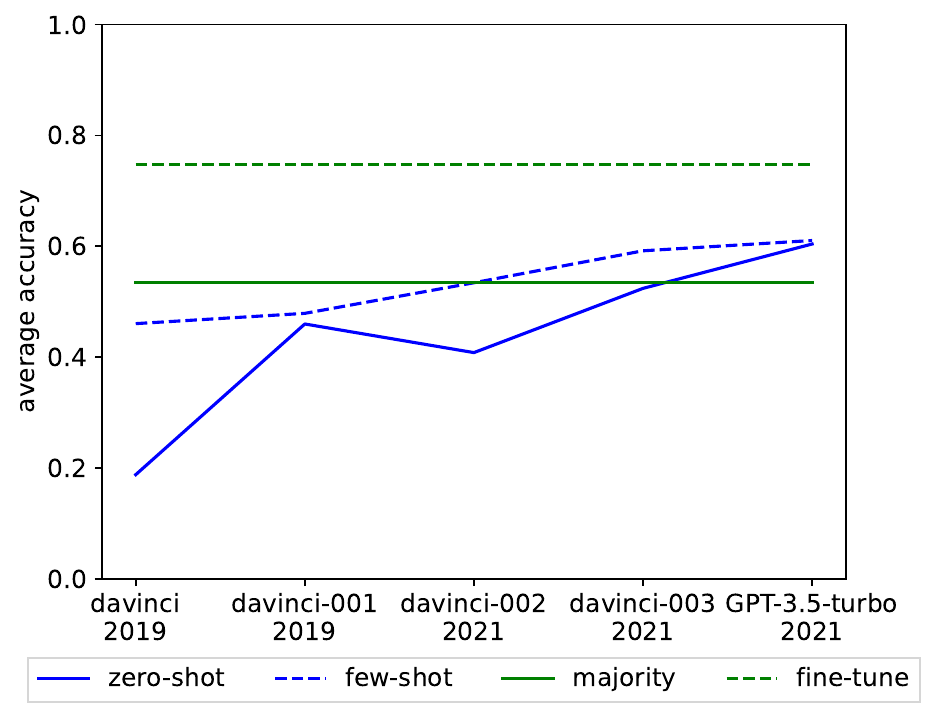}
            \caption[]%
            {{GPT-3 series on post-2021 datasets.}}    
            \label{fig:GPT-new-datasets}
        \end{subfigure}
        \begin{subfigure}[b]{0.475\textwidth}
            \centering
            \includegraphics[scale=0.45]{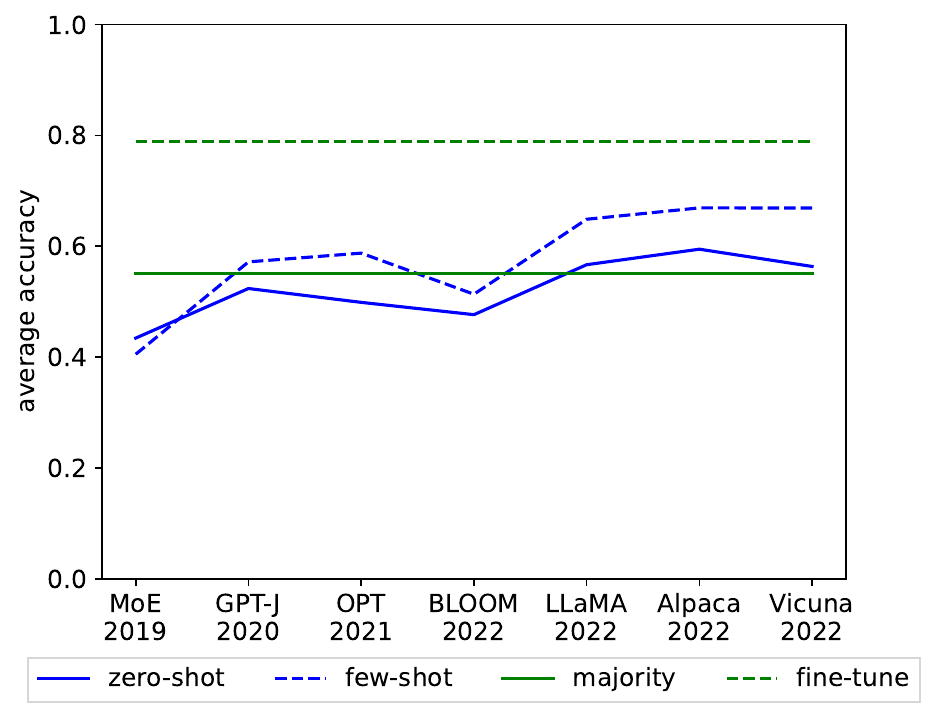}
            \caption[]%
            {{Open LLMs on pre-2021 datasets.}}    
            \label{fig:Open-old-datasets}
        \end{subfigure}
        \begin{subfigure}[b]{0.475\textwidth}   
            \centering 
            \includegraphics[scale=0.45]{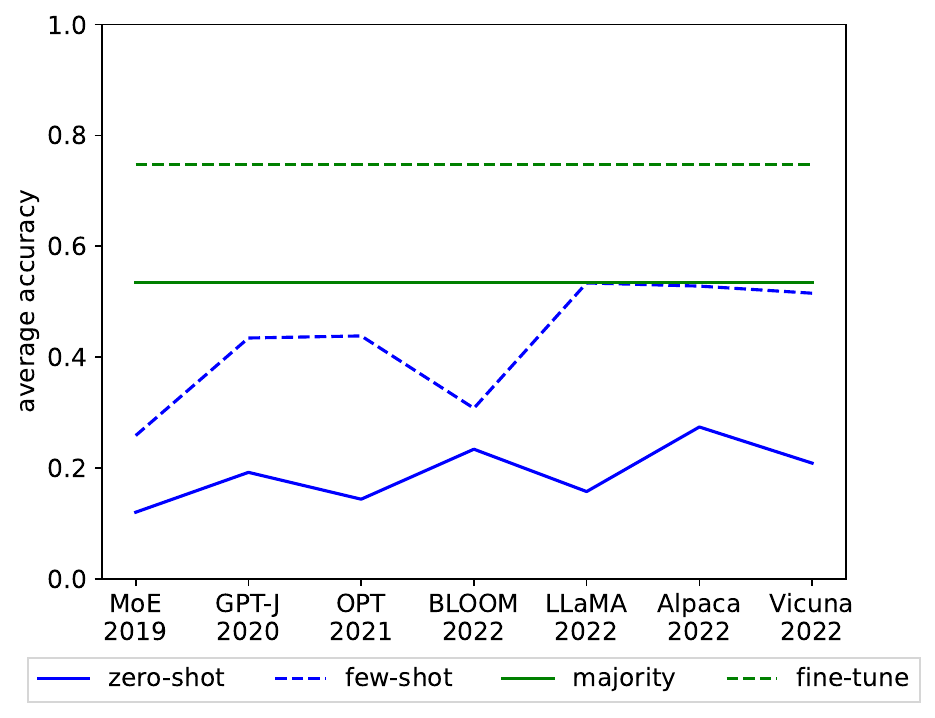}
            \caption[]%
            {{Open LLMs on post-2021 datasets.}}    
            \label{fig:Open-newdatasets}
        \end{subfigure}
        \caption[]
        {Average performance on datasets pre/post-2021.  In the $x$ axis, LLMs are ordered chronologically by training data collection date (collection year is listed below the LLM).
        } 
        \label{fig:experiments-average}
    \end{figure*}

\section{Chronological Analysis}
\label{sec:chronological}
We start with a chronological analysis. This allows us to detect patterns of possible task contamination across the LLMs and datasets we examine.

\subsection{Analysis of Pre- and Post-collection Datasets}

We perform a global chronological analysis across all datasets and LLMs. We look at the difference between performance on datasets released before the training data collection date for the LLM (\textbf{pre-collection}) versus after the training data collection date (\textbf{post-collection}). Specifically, we focus on whether the model is above the majority baseline.\footnote{The majority baseline for a classification task is the performance of a model that labels every example with the label that occurs most frequently in the dataset.} In this section we use this measure, instead of averaging the performance across datasets, to avoid datasets with large performance differences dominating the analysis.

With 12 models and 16 datasets, we have 192 model/dataset combinations.  Of these combinations, 136 the datasets were released before the LLM training data collection date (pre-collection) and 56 the dataset were release after (post-collection). For both sets, we compute the percentage of model/dataset combinations for which the model beats the majority baseline, both zero-shot and few-shot. The results are shown in Fig.~\ref{fig:baseline-percentage}. We find that for datasets released prior to the creation of the LLM, it is more likely the LLM beats the majority baseline for both zero and few-shot settings. Using the Mann-Whitney U test \cite{Mann1947}, we find the difference in those above the majority baseline between pre- and post-collection populations to be statistically significant at the 99\% confidence level for both zero and few shot settings. \nilay{This can be clearer. I don't understand exactly what the M-W test showed here. What are the two compared populations? Is this saying "of the models which outperform the majority baseline, the performance on the pre-collection dataset is significantly higher than on the post-collection datasets"? If so, why only consider the cases which outperform the baseline?}

For some model/dataset combinations, the performance difference above the majority baseline is small, so we also we compute the percentage of model/dataset combinations and for which the model beats the majority baseline and the difference above the majority baseline is statistically significant at the 99\% level\nilay{I don't quite understand what this means. How is this different than the previous test?}, calculated using the student t-test \cite{student1908} (Fig.~\ref{fig:baseline-percentage}, darker).  Again, we find that for datasets released prior to the creation of the LLM, it is far more likely the LLM beats the majority baseline with statistical significance for both zero and few-shot settings. Similarly, the Mann-Whitney U test indicates these differences between pre and post are statistically significant at the 99\% confidence level for both zero and few shot settings\nilay{I think the previous few paragraphs can be slightly reworded to better explain exactly what the statistical tests are demonstrating. I'll need a little time to think about how I would word it so I'll come back to this later.}.

These results indicate the possibility of task contamination for open LLMs and GPT-3 series LLMs.

\jmfb{add the caveats to this section (FOMC dataset, and that performance difference doesn't indicate )}

\subsubsection{Caveats}

\jmf{talk about FOMC dataset: that they used GPT-3 series, and it seems it may be contaminated}

There are two considerations we need to make in the global chronological analysis. 

\begin{figure}[t!]
          \centering
            \includegraphics[scale=0.35]{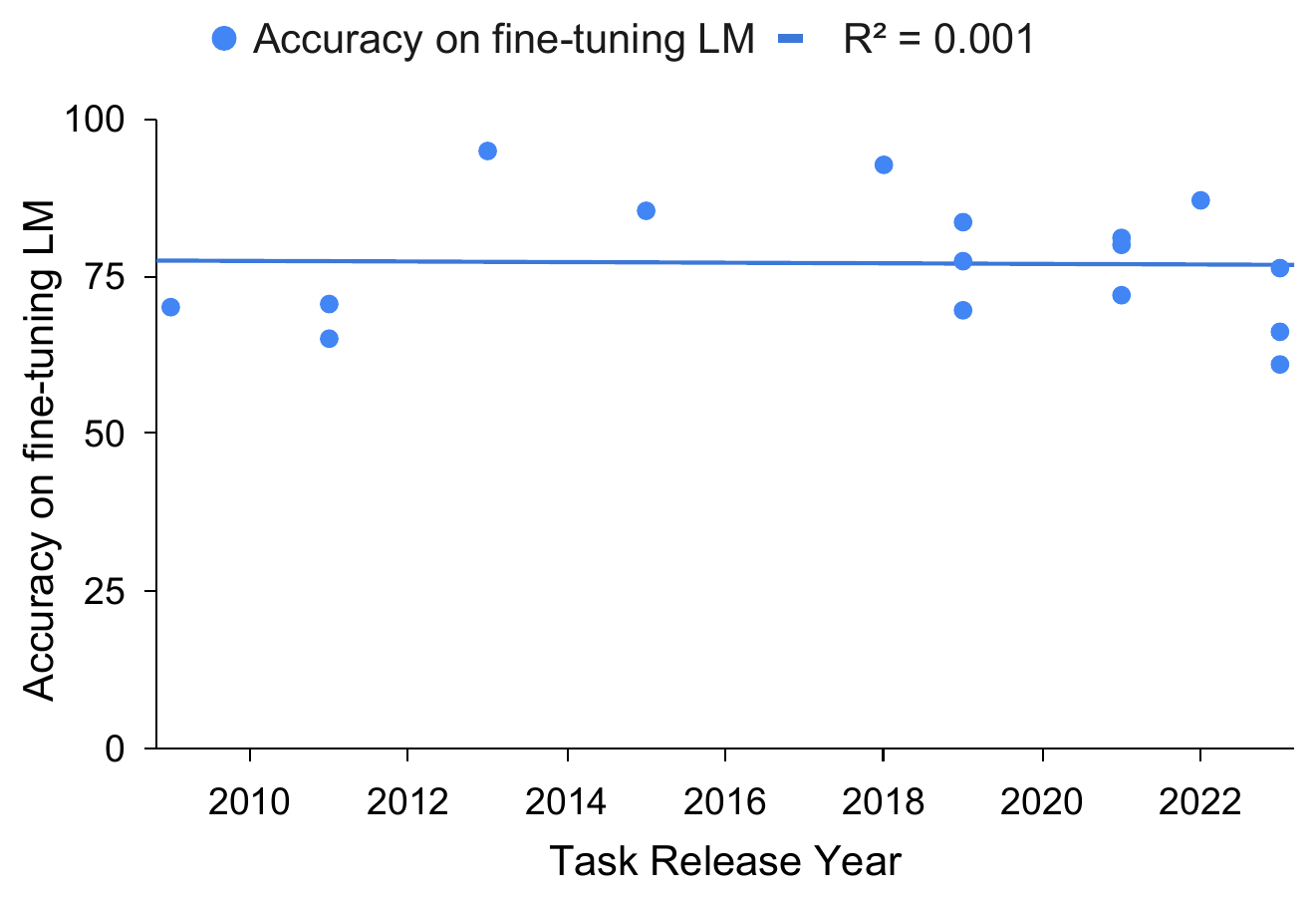}
    \caption[]%
            {Task accuracy of a fine-tuned LLM baseline vs. task release year. $R^2 = .001$, which indicates that the task difficulty for our datasets does not increase over time. \jmf{compute average performance on pre-2021 and post-2021, and also statistical significance.} \cm{the average score of pre-2021:78.82, post-2021:74.81. Student t-sest: the p-value is 0.221667. The result is not significant at p < 0.05.} }
            \label{fig:fine-tune-on-llm}
\end{figure}

First, datasets may have become more difficult over time, meaning LLMs are less likely to outperform the majority baseline despite the lack of task contamination. To account for this, we carefully review the tasks and remove tasks known to be difficult for LLMs, such as GSM8K \cite{Cobbe2021} and TrackingShuffledObjects \cite{Srivastava2023}. The remaining datasets all have acceptable performance using fine-tuned pretrained language models (PLMs), and, importantly, there is no correlation between release date and the performance of fine-tuned PLMs ($R^2 = 0.001$) on our datasets, as shown in Fig.~\ref{fig:fine-tune-on-llm}.

Secondly, post-collection datasets, despite being released after data collection, may still suffer from contamination. For example, the FOMC dataset \cite{shah2023} was officially released post-collection for the GPT-3 series, but the performance of subsequent versions of GPT-3 is notably high. This may be the result of the authors' preliminary experimentation with the GPT-3 series (as stated in their paper), as OpenAI may have then utilized their experimental data for model updates.

\subsection{Analysis of Pre- and Post-collection for Individual LLMs}


In this section, we consider the performance on pre- and post-collection datasets for each LLM individually (see Fig.~\ref{fig:experiments-majority}).\jmf{Add summary of results. We find...} We find the difference in performance between the two categories to be statistically significant at 95\% confidence according to the paired sign test \cite{Dixon1946}.

We plot the percentage of datasets larger than the majority baseline as in the last section, but for each LLM individually. The results are shown in Fig.~\ref{fig:experiments-majority}. We observe that the global trend from the previous section has remained true across models with the full range of dates, further indicating that the absolute date of the dataset is not the main factor, but rather the date of the dataset relative to the training data collection date for the LLM is the more important factor. (Note: because of the recency of BLOOM, LLaMA, Alpaca, and Vicuna, we have fewer datasets in our experiments post their training data collection date).  The results indicate the possibility of task contamination for both open LLMs and GPT-3 series LLMs, with a stronger indication of contamination in the GPT-3 series with \texttt{davinci-001} and after.

\subsection{Performance over Time}

Next we perform a chronological analysis that examines the change in average performance over time for both GPT-3 series and open LLMs (Fig.~\ref{fig:experiments-average}).\jmf{add summary of results} In the $x$ axis, LLMs are ordered chronologically by training data collection date. To also be sensitive to time of the datasets, we split our datasets into two sets: 
datasets released before or after January 1st, 2021, identified as \textbf{pre-2021} datasets and \textbf{post-2021} datasets, respectively.

\paragraph{Pre-2021 Datasets}
For open LLMs, on pre-2021 datasets, we see a slight increase over time for open LLMs (Fig.~\ref{fig:Open-old-datasets}).  We find that the performance hovers around the majority baseline for both zero and few-shot settings, and does not increase very much from LLM data collection dates ranging from 2019 to 2022.

For the GPT-3 series, on the other hand, the trend on pre-2021 datasets is particularly suspect (Fig.~\ref{fig:GPT-old-datasets}).  We see that for prior GPT-3 datasets, the performance has increased dramatically over time, with later \texttt{davinci} models much higher than the majority baseline for both zero and few-shot settings.  The comparison to open LLMs indicates that zero and few-shot evaluations may have task contamination issues due to data collected from user inputs.

\paragraph{Post-2021 Datasets}
For post-2021 datasets, GPT-3 average performance has also increased over time (Fig.~\ref{fig:GPT-new-datasets}), particularly in the zero-shot setting.  This makes sense, as many of the post-2021 datasets are released prior the training data collection date for the later \texttt{davinci} models.  (To see which datasets are pre- or post- training data collection time, see the line separating pre- and post- collection datasets in Table~\ref{tab:extract}.)  Open LLMs average performance also increased over time, but they remain lower than the majority baseline and the GPT-3 series.

One could hypothesize that the high performance of the GPT-3 series is due to instruction tuning \cite{ouyang2022}, however we do not believe this is the case.  While we observe an increase in performance from \texttt{davinci-001} to \texttt{davinci-002} on pre-2021 datasets, there is a corresponding decrease in performance on post-2021 datasets, which we measure with the sign test to be statistically significant at the 95\%\cm{Increase on zero-shot pre p\_value=0.00408, Decrease on zero-shot post p\_value=0.02939.}. This demonstrates that the GPT-3 series instruction tuning is specific to certain earlier datasets, and suggests dataset contamination for zero and few-shot evaluation of GPT-3 series.

\section{Training Data Inspection}
\label{sec:trainingdatainspection}
To search for direct evidence of task contamination, we conduct training data inspection on two instruction fine-tuned open LLMs (Alpaca and Vicuna) for all experimented classification tasks. We search for task-related instruction patterns in the training data, and manually inspect them to see if they contain task training examples. Because we must check manually, we can perform this analysis only for the small fine-tuning datasets of Alpaca and Vicuna. We then compare the performance to see if more task-specific training examples has boosted performance.

Table~\ref{tab:open-source-llm-contam} shows the number of task examples on Alpaca and Vicuna, as well as the change in performance over LLaMA averaged over zero and few-shot settings and all tasks.  We find that performance has improved for Alpaca and Vicuna over the original LLaMA model for tasks with more than one task example. Because Alpaca and Vicuna are fine-tuned LLaMA models, this indicates that the performance can be improved with small sets of task examples in the training data, which can compromise zero-shot or few-shot evaluation. \nilay{This could also be the effect of instruction tuning. If possible, can we maybe re-train alpaca minus the examples you extracted and see how it does? If it's significantly worse, that is very strong evidence for task contamination. I think training alpaca is feasible in a couple days on Nautilus, and is easy to run since code is provided. Maybe could try Alpaca-LoRA if it's too heavy.}

\begin{table}[t!]
\centering \small
\begin{tabular}{l||lll}
Dataset &  Alpaca & Vicuna  \\ \hline \hline
RTE  &    0, +3.1\%   &\textbf{33, +10.6\%} \\
WNLI &   0, -1.4\%   &\textbf{33, +7.7\%} \\
COPA &   ?, 0\%    & ?, +10\%\\
SST-2 &  \textbf{8, +14.6\%}  & 0, -1.0\% \\
MRPC &  0, -0.7\%  & 0, -8.0\% \\
QNLI &   0, -0.4\%   &\textbf{28, +10.0\%} \\
CB   &   0, +9.8\%   &0, -23.2\%\\
WiC &    0, -4.9\%   & 0, -2.5\% \\
BoolQ &  ?, +1.9\%    & ?, +4.0\%\\
\hline \hline
StrategyQA   & 0, -3.3\%   & 0, +10.3\% \\
MTSC-RW  &  ?,  +9.6\%& ?, +11.3\% \\
MTSC-MT  &  ?, +6.9\%& ?, +8.0\% \\
NLI4Wills    & 0, -13.5\%    & 0, -11.6\%  \\
CREPE  & 0, +24.2\%  & 0, -0.4\%\\
FOMC  & 0, -5.7\%   & \textbf{1, -5.4\%}\\
NewsMet  & \textbf{4, +7.2\%}   & 0, -11.4\%\\
\end{tabular}
\caption[]{Training data inspection results: \# of datapoints in the Alpaca and Vicuna datasets that are examples of the task, and $\Delta$\%, the performance difference compared to LLaMA averaged across zero and few-shot settings. Task examples are found by matching a regular expression for the task followed by a manual inspection. Bold indicates task examples are found.  "?" indicates there is no specific pattern to match, so we cannot count the number of examples.  Regular expressions for each task are listed in the Appendix \ref{app:inspection}.} 
\label{tab:open-source-llm-contam}
\end{table}

\begin{table*}[]
\centering\resizebox{\textwidth}{!}{
\begin{tabular}{lllllllllllll}
\toprule
Task & Davinci                    & davinci-001           & davinci-002           & davinci-003           & GPT-3.5-T             & MoE           & GPT-J                   & OPT                     & Bloom                   & LLaMA                   & Alpaca                  & Vicuna                 \\
\midrule
RTE        & \grysq & \redx    & \redx    & \redx    & \redx   & \grysq & \grysq & \grysq & \grysq & \grysq & \grnsq & \redx    \\
WNLI       & \grysq & \redx    & \redx    & \redx    & \redx    & \grysq & \grysq & \grysq & \grysq & \grysq & \grnsq & \redx    \\
COPA      & \grysq & \grnsq & \grnsq & \redx    & \redx    & \grysq & \grysq & \grysq & \grysq & \grysq & \grnsq & \grnsq \\
SST-2      & \grysq & \grnsq & \redx    & \redx    & \redx    & \grysq & \grysq & \grysq & \grysq & \grysq & \grnsq & \grnsq \\
MRPC       & \grysq & \grnsq & \grnsq & \redx    & \redx    & \grysq & \grysq & \grysq & \grysq & \grysq & \grnsq & \grnsq \\
QNLI       & \grysq & \grnsq & \redx    & \redx    & \redx    & \grysq & \grysq & \grysq & \grysq & \grysq & \grnsq & \grnsq \\
CB        & \grysq & \redx    & \redx    & \redx    & \redx   & \grysq & \grysq & \grysq & \grysq & \grysq & \grnsq & \grnsq \\
WiC       & \grysq & \grnsq & \redx    & \redx    & \redx    & \grysq & \grysq & \grysq & \grysq & \grysq & \grnsq & \grnsq \\
BoolQ     & \grysq & \grnsq & \grnsq & \redx    & \redx    & \grysq & \grysq & \grysq & \grysq & \grysq & \grnsq & \grnsq \\
\cline{1-3} \cline{7-8}
StrategyQA & \grysq & \grnsq & \multicolumn{1}{|l} \grnsq & \grnsq & \grnsq & \multicolumn{1}{|l} \grysq & \grysq & \multicolumn{1}{|l} \grysq & \grysq & \grysq & \grnsq & \grnsq \\
NewsMTSC-MT  & \grysq & \grnsq & \multicolumn{1}{|l} \grnsq & \redx    & \redx    & \multicolumn{1}{|l} \grysq & \grysq & \multicolumn{1}{|l} \grysq & \grysq & \grysq & \grnsq & \redx    \\
NewsMTSC-RW  & \grysq & \grnsq & \multicolumn{1}{|l} \grnsq & \redx    & \redx    & \multicolumn{1}{|l} \grysq & \grysq & \multicolumn{1}{|l} \grysq & \grysq & \grysq & \grnsq & \redx    \\
\cline{4-6} \cline{9-9}
NLI4Wills & \grysq & \grnsq & \grnsq & \grnsq & \grnsq & \grysq & \grysq & \grysq & \multicolumn{1}{|l} \grysq & \grysq & \grnsq & \grnsq \\
\cline{10-13}
CREPE     & \grysq & \grnsq & \grnsq & \grnsq & \grnsq & \grysq & \grysq & \grysq & \grysq & \grysq & \grnsq & \grnsq \\
FOMC      & \grysq & \grnsq & \redx    & \redx    & \redx    & \grysq & \grysq & \grysq & \grysq & \grysq & \grnsq & \grnsq \\
NewsMet    & \grysq & \grnsq & \grnsq & \redx    & \redx    & \grysq & \grysq & \grysq & \grysq & \grysq & \grnsq & \grnsq \\
\bottomrule
\end{tabular}}
\caption{Task example extraction results on all tasks (tasks ordered top to bottom by release date). A line separates those datasets released before the LLM's training data collection date (pre-collection, top) and those after (post-collection, bottom) for each LLM. \redx~indicates the model can generate training examples for the task. We indicate models with instruction tuning and those without using \grnsq~and \grysq, respectively.  \grnsq~indicates a model with instruction tuning cannot generate task examples, while \grysq~indicates a model without instruction tuning cannot generate task examples.  Models without instruction tuning cannot follow the instructions directing them to generate task examples.\jmfb{possibly confusing so changed to green square (\grnsq) vs red X (\redx)} \jmf{can we do an analysis correlating task contamination with above majority baseline for all models?}\jmf{Important: non-instruction tuned models don't know instructions, so they can't seem to generate training examples}}
\label{tab:extract}
\end{table*}

\section{Task Example Extraction} \jmf{rename to Task Example Extraction?}
\label{sec:trainingdataextraction}
We test for task data contamination by attempting to extract task examples from the LLM.  Prior work~\cite{lm-contamination} has tested if there exists testing data contamination by prompting an LLM to generate examples for a task.  If the LLM can generate examples that exactly match examples in the test data, it is evidence that the test set of the task has been seen during training by the LLM. Inspired by their method, we adopt a similar approach to test for task contamination. Instead of attempting to generate test data, we prompt the model to generate training examples, since for zero- or few-shot evaluation, the model should not be trained on any task examples. If an LLM can generate training examples based on the prompt, this is evidence of task contamination.  Note we do not require an exact match of the generated examples with the training data for the task, since any examples for the task seen during training indicate possible task contamination.  Our prompts for task example extraction are given in Appendix~\ref{app:prompt_extraction}.

Table \ref{tab:extract} shows the task example extraction results on all tasks across all models. For all \textbf{pre-collection datasets}, GPT-3 series models starting from \texttt{davinci-001} can generate task specific training examples. There are some \textbf{post-collection datasets} that have evidence of contamination for the GPT-3 series. These datasets may have been contaminated if the authors of these datasets experimented with the GPT-3 series before releasing the dataset. For example, the FOMC paper \cite{shah2023} states they tested with the GPT-3 series, which could have caused contamination. For open LLMs, almost no models can generate training examples of specific tasks except for Vicuna, which is fine-tuned on the ChatGPT data. Note models without instruction tuning cannot follow the instructions directing them to generate task examples, so this analysis is not conclusive for these models.

\subsection{Comparison to Training Data Inspection}

\jmf{Add Comparison to Training Data Inspection}

\jmf{Neither are perfectly reliable}

Comparing Tables~\ref{tab:open-source-llm-contam} and \ref{tab:extract}, we find that training data inspection (TDI) and task example extraction (TEE) both suffer from low recall.  TDI has demonstrated task contamination in Alpaca for SST-2 and NewsMet datasets, but TEE failed to catch this contamination.  Similarly, TEE has demonstrated task contamination for Vicuna for NewsMTSC, but TDI has failed to catch it.  Both suffer from low recall, and highlight the difficulties of employing these methods for detecting task contamination.

\begin{figure*}[t!]
          \centering
          \begin{subfigure}[b]{0.475\textwidth}
            \includegraphics[scale=0.5]{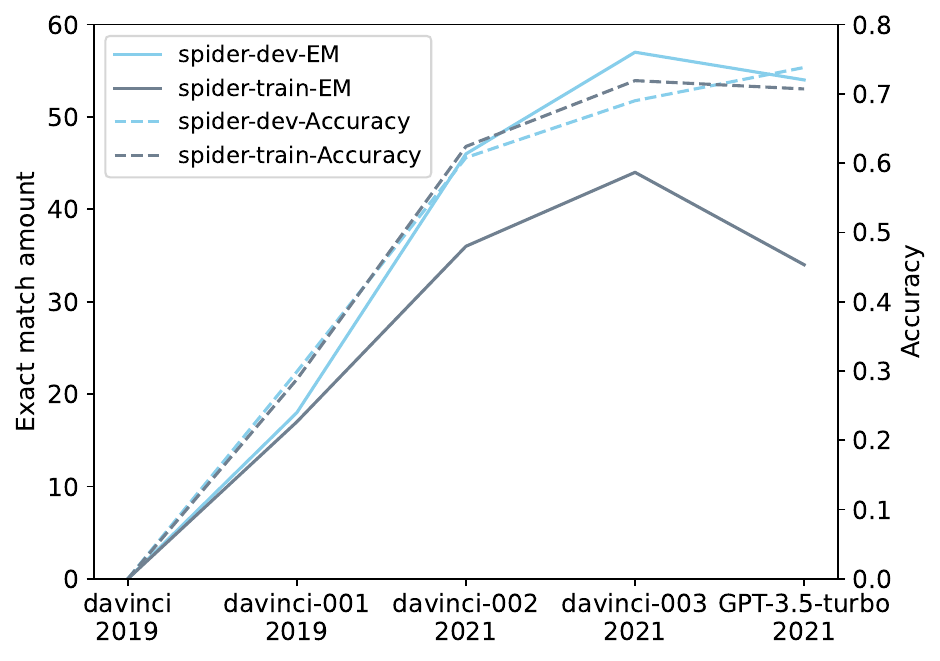}
            \caption[]%
            {{Over GPT-3 series.}}
            \label{fig:spider}
            \end{subfigure}
        \hfill
        \centering
        \begin{subfigure}[b]{0.475\textwidth}
            \includegraphics[scale=0.5]{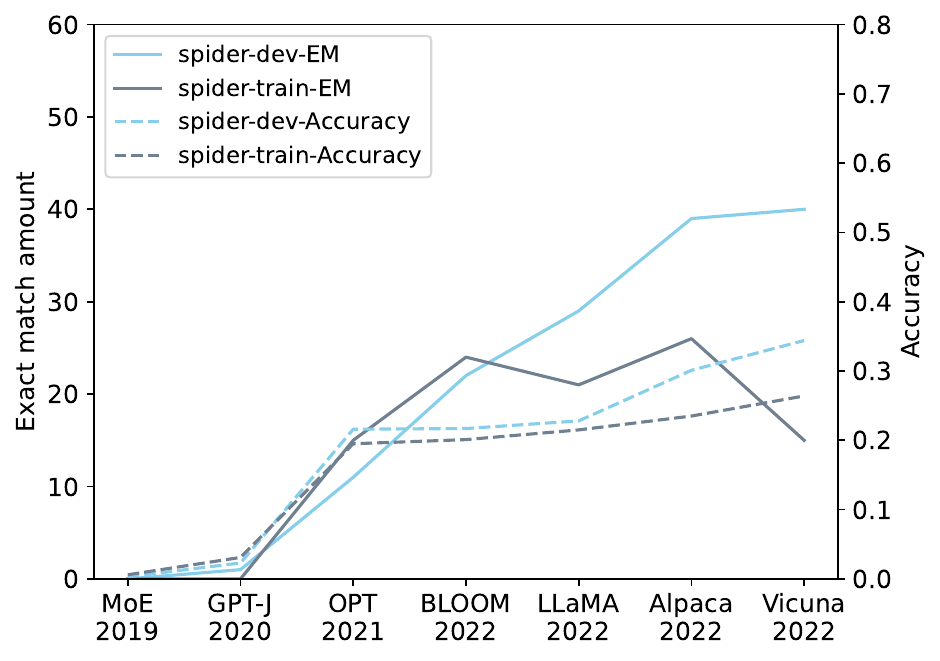}
            \caption[]%
            {{Over recent LLMs.\jmfb{Let's try to explain why Vicuna drops on dev-EM, and not on train-EM, also when compared to Alpaca.}}}
            \label{fig:spider-2}
        \end{subfigure}
    \caption[]%
            {{The number of generated examples which exactly match the original set and the performance (accuracy).\jmfb{is this with schema, or without schema?,\cm{EM scores without schema, accuracy scores with schema}}}}
            \label{fig:spider-both}
\end{figure*}

\begin{figure*}[t!]
          \centering
            \includegraphics[scale=0.5]{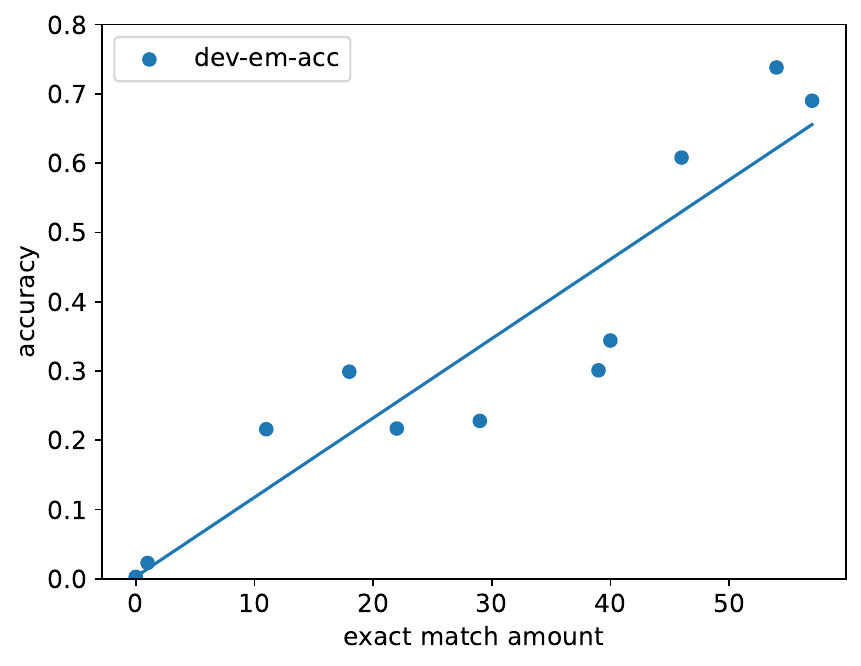}
    \caption[]%
            {Membership inference: Exact match count vs. accuracy for Spider on development set. $R^2=0.88$}
            \label{fig:dev-acc-em}
\end{figure*}

\section{LLM Performance on Tasks With No Contamination}
\label{sec:no_contamination}

We find that for tasks without demonstrated possibility of task contamination, LLMs rarely show statistically significant improvements over majority baselines. In Table~\ref{tab:extract}, for the $51$ model/dataset combinations that are post-collection and have no extracted task examples, only 1 out of 51, or $2\%$, demonstrate a statistically significant improvements over the majority baseline for either zero or few-shot settings.  This combination is \texttt{davinci-001} on MTSC-RW, which shows a statistically significant improvement over the majority baseline (Tables~\ref{tab:zero-shot} and \ref{tab:few-shot} in the Appendix) but does not generate task examples with our prompt.  This dataset is found by cross-referencing Table~\ref{tab:extract} and Tables~\ref{tab:zero-shot} and \ref{tab:few-shot} in the Appendix, and looking for datasets which are post-collection and not marked \redx~in Table~\ref{tab:extract}, and are bold in either Table~\ref{tab:zero-shot} or \ref{tab:few-shot}.

\section{Membership Inference}
\label{sec:membershipinference}
\jmf{This is not possible to do for classification tasks.  So we do it for a generation task.}

To further examine the effect of training data contamination, we apply a membership inference attack \cite{hu2022}, which checks if model generated content exactly matches the examples in the dataset. While this test is possible for generation tasks, it is not possible for classification tasks, since inputs may be in the training data of LLMs (and likely are, for many datasets), but we do not know for certain if the inputs are also paired with the labels without looking at the training data. We use Spider, a semantic parsing and text-to-SQL generation task, \cite{yu2018} as our target for analysis.

Fig.~\ref{fig:spider} and Fig.~\ref{fig:spider-2} show how many generated examples from the sampled training set and full development set are exactly the same over versions of the GPT-3 series and recent open sourced LLMs, respectively. The database schemas are not in the zero-shot prompts, so if the model can generate exactly the same table name or field name as found in the training or development data, there must be contamination. As shown in Fig.~\ref{fig:spider-both}, the number of exact matched generated examples increases over time, which indicates the extent of the task contamination on Spider is increasing. 

We also compute the execution accuracy after adding the schema in the prompts, and plot it against the number of exact matched generations (Fig.~\ref{fig:dev-acc-em}). We find a strong positive correlation between the number of exact matched generated examples and execution accuracy ($R = 0.88$), strongly indicating increased contamination is related to increased performance. However, we still cannot determine the extent of the contamination's effect on performance improvement. We leave this for future work.

\section{Take-Aways}
We now share some takeaways which our experiments have brought to light:

\begin{itemize}
    \item Due to task contamination, closed-sourced models may demonstrate inflated performance in zero-shot or few-shot evaluation, and are therefore not trustworthy baselines in these settings, especially those including instruction fine-tuning or reinforcement learning with human feedback (RLHF)\nilay{cite RLHF here?}. The extent of this contamination is still unknown, and we therefore recommend caution.

    \item In our experiments, for classification tasks without demonstrated possibility of task contamination, LLMs rarely show statistically significant improvements over majority baselines, in both zero and few-shot settings.

    \item The observed increase over time of GPT-3 series models for zero-shot or few-shot performance for many downstream tasks is likely due to task contamination. \nilay{kinda the same as the previous bullet. Think we can remove?}

    \item Inspection for task contamination of training data even for open-sourced LLMs can be difficult for several reasons. First, determining membership is difficult unless the processed dataset used for training the LLM is released (e.g., OPT and LLaMA did not release the data they used to train the model, but Alpaca and Vicuna did, so we can obtain more definite information). Second, we cannot always rely on the model to reproduce evidence of contamination even if it exists. And third, formatting differences (such as CSV and JSON) of a dataset complicate analysis.

    \item We encourage publicly releasing training datasets to allow for easier diagnosis of contamination issues. \nilay{Also related to previous bullet, we can probably find a way to merge these}
    
\end{itemize}

\jmfb{Recommend Against Using Highest Performing Closed Models as Baselines, since they may not be zero or few-shot.}

\jmfb{Performance increase over time for GPT-3, partially due to dataset contamination}

\jmfb{Contamination Analysis Difficulties with Web Data: Difficult to definitely decide if dataset is included unless we have the whole dataset (like OPT).  Still difficult without links.  Cannot always rely on model to reproduce data, since need to be paired with labels. Formatting can cause issues.}

\jmfb{Difficulties with Closed Dataset Models like GPT and ChatGPT that have been fine-tuned on dataset we don't know about.  They can have contamination issues.  Open models like Alpaca and Vicuna help to diagnose contamination issues.}

\section{Related Work}

The investigation into potential data contamination in large language models (LLMs) has recently been gaining attention in the research community. \citet{brown2020}, in their work with GPT-3, presented an in-depth analysis of data contamination. Although they acknowledged the presence of a bug that led to data contamination in multiple datasets, their position was that it did not affect the overall performance of the model. Intriguingly, they noted that contaminated datasets outperformed the uncontaminated ones which, in a way, contradicted their original assertion. \citet{magar2022} extracted training data from GPT-2 and indicated potential leaks of private data in the pre-trained language model. \citet{chang2023} discovered that OpenAI models were memorizing substantial amounts of copyrighted materials, which increased concern over data contamination. \citet{aiyappa2023} highlighted the severity and scope of data contamination problems for ChatGPT evaluations. Highlighting the need for strategic interventions to address these issues, \citet{jacovi2023} proposed several strategies for mitigating testing data contamination. Additional work has further looked into test data contamination \cite{lm-contamination, zhou2023, golchin2023, sainz2023, deng2023investigating, oren2023proving, li2023}.

The previous work listed above has investigated test data contamination, but has not considered task contamination for zero-shot or few-shot settings. Prior work has noticed our proposed task contamination problem for zero-shot or few-shot learning \cite{blevins2023, briakou2023}, but did not systematically analyze it. Our work seeks to add to the existing knowledge by providing an exhaustive evaluation of task contamination for few-shot or zero-shot learning scenarios. 


\section{Conclusion and Future Work}
We investigate task contamination for LLMs, and conduct a chronological analysis, training data inspection, task example extraction, and a membership inference attack to analyze it. We find evidence that some LLMs have seen task examples during pre-training for a range of tasks, and are therefore no longer zero or few-shot for these tasks.  Additionally, we find that for classification tasks with no possibility of task contamination, LLMs rarely demonstrate statistically significant improvements over simple majority baselines, in both zero and few-shot settings. We recommend additional research be conducted on task contamination for zero and few-shot settings to reveal the extent and impact of task contamination for large language models in these settings.

\section*{Acknowledgements}

We are grateful for valuable feedback from Nilay Patel on an earlier version of this draft.  
We are thankful for the computing resources provided by the Pacific Research Platform's Nautilus cluster, supported in part by National Science Foundation (NSF) awards CNS-1730158, ACI-1540112, ACI-1541349, OAC-1826967, OAC-2112167, CNS-2100237, CNS-2120019, the University of California Office of the President, and the University of California San Diego's California Institute for Telecommunications and Information Technology/Qualcomm Institute. Thanks to CENIC for the 100Gbps networks.


\bibliography{aaai24}

\newpage

\appendix

\section{Hyperparameters}
\label{app:hyperparameters}
We use greedy decoding to ensure a fair comparison for all approaches. For GPT-3 series models, we set the temperature as 0 to ensure deterministic results. For few-shot learning, we use the same few-shot examples across models for each instance in a task. We run open sourced models on an NVIDIA A100 GPU.

\section{Datasets}
\label{app:datasets}
\jmf{update}The pre-2021 datasets are common GLUE \cite{wang2018} and Super GLUE \cite{Wang2019} tasks: MRPC \cite{dolan-brockett-2005}, boolq \cite{clark2019}, SST-2 \cite{socher2013}, QNLI \cite{Demszky2018}, WNLI \cite{levesque2012}, RTE \cite{Giampiccolo2008}, CB \cite{Marneffe2019}, COPA \cite{roemmele2011}, WiC \cite{pilehvar2019}. The post-2021 datasets are StrategyQA \cite{geva2021},  NLI4Wills \cite{kwak2022}, NewsMTSC \cite{hamborg2021}, CREPE \cite{yu2023}, FOMC \cite{shah2023} and NewsMet \cite{joseph2023}.

\begin{table}[h!]
\centering \small
\begin{tabular}{l||ll}
Dataset & Year  & Test set size\\ \hline \hline
RTE & 2009  & 277 \\
WNLI & 2011 & 71\\
COPA & 2011  & 100\\
SST-2 & 2013 & 872\\
MRPC & 2015  & 408 \\
QNLI & 2018  & 5463\\
CB &  2019 & 56\\
WiC & 2019  & 638\\
BoolQ &  2019 & 3270\\
\hline \hline
StrategyQA &  2021  & 229\\
NewsMTSC-mt & 2021 & 1476\\
NewsMTSC-rw & 2021 & 1146\\
NLI4Wills & 2022 & 255\\
CREPE & 2023 & 2000 \\
FOMC & 2023 &496\\
NewsMet& 2023 &554

\end{tabular}
\caption[]{Dataset release year and test set size for each task.}
\label{tab-size-year-llm}
\end{table}

\section{Prompt Sources}
The prompts for these tasks are taken from previous research \cite{Bang2023, Qin2023} that use them as evaluation benchmarks and \citet{openai2023b} Examples or designed based on the related tasks from these sources. Table \ref{fig-prompt-source-llm} shows prompt source for each dataset.  Appendix~\ref{app:prompt_examples} lists example prompts for each task.

\begin{table}[h!]
\centering \small
\begin{tabular}{l||ll}
Dataset & Prompt source  \\ \hline \hline
RTE &  \citet{Bang2023}* \\
WNLI &  \citet{Bang2023}* \\
COPA &  \citet{Bang2023}*\\
SST-2 & \citet{openai2023b}\\
MRPC &  \citet{openai2023b}*  \\
QNLI &  \citet{Bang2023}* \\
CB &  \citet{Bang2023}*\\
WiC &  \citet{openai2023b}* \\
BoolQ &  \citet{Qin2023}* \\
\hline \hline
StrategyQA &   \citet{Qin2023} \\
Newsmtsc-mt & \citet{openai2023b}* \\
Newsmtsc-rw & \citet{openai2023b}* \\
NLI4Wills &  \citet{Bang2023}* \\
CREPE &   \citet{openai2023b}* \\
FOMC &  \citet{shah2023} \\
NewsMet& \citet{Bang2023}* 
\end{tabular}
\caption[]{Prompt source for each task. * indicates we designed our prompt based on the referenced source.}
\label{fig-prompt-source-llm}
\end{table}

\section{Training Data Inspection Details}
\label{app:inspection}
We manually inspect training examples found using regular expressions for each task. Our regular expression or string search pattern for each task are listed in Table~\ref{tab-re-patterns}. Some tasks such as COPA and BoolQ do not have a specific pattern that can be matched. We count an example if it is directly related to the task and contains the input and output for the task.  We do not count examples that talk about the task without giving input and output examples.

\begin{table}[t!]
\centering \small
\begin{tabular}{l||l}
Dataset & RE pattern  \\ \hline \hline
RTE  &  [Ee]ntailment \\
WNLI &  [Ee]ntailment \\
COPA &  -- \\
SST-2 &  [cC]lassify the sentiment\\
MRPC &  [Pp]paraphrase \\
QNLI &  [Ee]ntailment \\
CB   &  [Ee]ntailment \\
WiC &   [Ww]ord sense \\
BoolQ & -- \\
\hline \hline
StrategyQA &  ([tT]he answer is)*([Yy]es|[Nn]o) \\
NLI4Wills &  [sS]upport|[Rr]efute\\
MTSC-RW & --\\
MTSC-MT & --\\
CREPE & presupposition \\
FOMC & "hawkish" or "dovish"\\
NewsMet & "metaphorical" \\
\end{tabular}
\caption[]{RE patterns used for each task. -- indicates there is no specific pattern to match for this task.}
\label{tab-re-patterns}
\end{table}

\onecolumn

\section{Detailed Results Tables}
\label{app:detailed_results}

In this section, we report the performance numbers for all models and datasets in our experiments with confidence intervals.

\begin{table*}[h!]
\centering\resizebox{\textwidth}{!}{
\begin{tabular}{l||l|lllll||lllllll}
Dataset                   & Majority            & davinci & davinci-001 & davinci-002 & davinci-003 & GPT-3.5-T & MoE-7B & GPT-J-6B & OPT-6.7B  & BLOOM-7B                       & LLama-7B & Alpaca-7B & Vicuna-7B  \\ \hline \hline
        RTE &     52.7 &  29.6$\pm$2.9 &           57.4$\pm$3.5 &  \textbf{75.1$\pm$2.6} &  \textbf{83.8$\pm$1.9} &  \textbf{72.6$\pm$2.8} &           61.7$\pm$3.3 &  53.1$\pm$3.5 &  53.1$\pm$3.5 &           52.7$\pm$3.5 &           63.2$\pm$3.3 &           54.9$\pm$3.5 &           60.7$\pm$3.4 \\
       WNLI &     56.3 &  33.8$\pm$6.4 &           43.7$\pm$7.0 &           66.2$\pm$6.4 &           60.6$\pm$6.8 &           66.2$\pm$6.4 &           45.1$\pm$7.1 &  43.7$\pm$7.0 &  43.7$\pm$7.0 &           43.7$\pm$7.0 &           46.5$\pm$7.1 &           43.7$\pm$7.0 &           43.7$\pm$7.0 \\
       COPA &     55.0 &  66.0$\pm$5.4 &           70.0$\pm$5.0 &  \textbf{89.0$\pm$2.3} &  \textbf{93.0$\pm$1.6} &  \textbf{82.0$\pm$3.5} &           56.0$\pm$5.9 &  50.0$\pm$6.0 &  53.0$\pm$5.9 &           53.0$\pm$5.9 &           55.0$\pm$5.9 &           58.0$\pm$5.8 &           72.0$\pm$4.8 \\
      SST-2 &     50.9 &   0.3$\pm$0.0 &           58.0$\pm$1.9 &  \textbf{85.1$\pm$1.0} &  \textbf{73.4$\pm$1.5} &  \textbf{81.8$\pm$1.2} &            5.4$\pm$0.4 &  49.1$\pm$2.0 &  34.7$\pm$1.8 &           53.4$\pm$2.0 &           57.8$\pm$1.9 &  \textbf{87.3$\pm$0.9} &  \textbf{62.0$\pm$1.9} \\
       MRPC &     68.4 &   9.3$\pm$1.0 &           68.4$\pm$2.5 &           68.4$\pm$2.5 &           72.5$\pm$2.3 &           69.9$\pm$2.4 &           34.8$\pm$2.6 &  69.9$\pm$2.4 &  55.6$\pm$2.9 &           31.6$\pm$2.5 &           68.9$\pm$2.5 &           68.4$\pm$2.5 &           68.4$\pm$2.5 \\
       QNLI &     50.5 &  28.0$\pm$0.6 &           49.5$\pm$0.8 &  \textbf{57.2$\pm$0.8} &  \textbf{84.6$\pm$0.4} &  \textbf{85.1$\pm$0.4} &  \textbf{55.0$\pm$0.8} &  49.7$\pm$0.8 &  53.0$\pm$0.8 &           49.5$\pm$0.8 &           51.5$\pm$0.8 &           49.6$\pm$0.8 &  \textbf{59.0$\pm$0.8} \\
         CB &     50.0 &  35.7$\pm$7.5 &           75.0$\pm$6.1 &           75.0$\pm$6.1 &           76.8$\pm$5.8 &           75.0$\pm$6.1 &           26.8$\pm$6.4 &  44.6$\pm$8.1 &  41.1$\pm$7.9 &           50.0$\pm$8.1 &           41.1$\pm$7.9 &           48.2$\pm$8.1 &           12.5$\pm$3.6 \\
        WiC &     50.0 &  16.3$\pm$1.2 &           45.5$\pm$2.2 &           48.9$\pm$2.2 &  \textbf{60.5$\pm$2.1} &           54.4$\pm$2.2 &           50.3$\pm$2.2 &  51.3$\pm$2.2 &  55.3$\pm$2.2 &           50.5$\pm$2.2 &  \textbf{59.6$\pm$2.2} &           50.3$\pm$2.2 &           52.7$\pm$2.2 \\
      BoolQ &     62.2 &  19.6$\pm$0.6 &  \textbf{78.7$\pm$0.6} &  \textbf{83.5$\pm$0.5} &  \textbf{85.0$\pm$0.5} &  \textbf{87.1$\pm$0.4} &           55.8$\pm$0.9 &  60.1$\pm$0.9 &  59.5$\pm$0.9 &           44.6$\pm$0.9 &  \textbf{66.5$\pm$0.8} &  \textbf{74.9$\pm$0.7} &  \textbf{76.3$\pm$0.7} \\\cline{1-1} \cline{2-4} \cline{8-9}
 StrategyQA &     53.3 &  31.9$\pm$3.4 &           55.9$\pm$3.8 &           \multicolumn{1}{|l}{53.7$\pm$3.9} &           62.0$\pm$3.7 &           65.1$\pm$3.5 &           46.7$\pm$3.9 &  23.6$\pm$2.8 &  \multicolumn{1}{|l}{12.2$\pm$1.7} &           24.0$\pm$2.8 &           36.2$\pm$3.6 &           21.8$\pm$2.7 &           53.3$\pm$3.9 \\
    MTSC-MT &     50.7 &   3.3$\pm$0.2 &           48.8$\pm$1.5 &           \multicolumn{1}{|l}{34.8$\pm$1.4} &  \textbf{63.8$\pm$1.4} &  \textbf{67.1$\pm$1.3} &            0.0$\pm$0.0 &   4.2$\pm$0.2 &   \multicolumn{1}{|l}{2.6$\pm$0.2} &            3.3$\pm$0.2 &            2.2$\pm$0.1 &            5.1$\pm$0.3 &           12.3$\pm$0.7 \\
    MTSC-RW &     39.7 &   4.5$\pm$0.3 &  \textbf{50.4$\pm$1.7} &           \multicolumn{1}{|l}{34.8$\pm$1.6} &  \textbf{60.9$\pm$1.6} &  \textbf{69.2$\pm$1.5} &            0.0$\pm$0.0 &   4.3$\pm$0.3 &   \multicolumn{1}{|l}{3.1$\pm$0.2} &            3.3$\pm$0.2 &            2.3$\pm$0.2 &            7.8$\pm$0.5 &           10.7$\pm$0.7 \\\cline{5-7} \cline{10-10}
  NLI4Wills &     55.7 &  17.6$\pm$2.1 &           23.1$\pm$2.6 &           15.7$\pm$1.9 &           33.7$\pm$3.3 &           41.6$\pm$3.6 &           14.5$\pm$1.8 &  14.5$\pm$1.8 &   2.0$\pm$0.3 &            \multicolumn{1}{|l}{3.5$\pm$0.5} &            7.1$\pm$1.0 &           19.2$\pm$2.3 &           21.6$\pm$2.5 \\ \cline{11-14}
      CREPE &     72.8 &  20.5$\pm$0.9 &           40.1$\pm$1.3 &           28.1$\pm$1.1 &           42.1$\pm$1.3 &           69.3$\pm$1.1 &            4.1$\pm$0.2 &  16.5$\pm$0.7 &  44.3$\pm$1.3 &           68.5$\pm$1.1 &           20.4$\pm$0.8 &           67.2$\pm$1.1 &           18.1$\pm$0.8 \\
       FOMC &     49.4 &  33.3$\pm$2.3 &           52.6$\pm$2.6 &  \textbf{61.5$\pm$2.5} &           54.0$\pm$2.6 &           59.5$\pm$2.5 &           11.1$\pm$1.0 &  24.2$\pm$1.9 & 11.5$\pm$1.1 & 25.0$\pm$2.0 & 39.1$\pm$2.5 & 25.0$\pm$2.0 & 28.4$\pm$2.1 \\
    NewsMet &     52.3 &  20.4$\pm$1.6 &           50.9$\pm$2.5 &           57.0$\pm$2.4 &           50.2$\pm$2.5 &           51.1$\pm$2.5 &            7.8$\pm$0.7 &  47.5$\pm$2.5 &  34.8$\pm$2.3 &           36.1$\pm$2.3 &           31.0$\pm$2.1 &           46.9$\pm$2.5 &            8.7$\pm$0.8 \\
\end{tabular}}
\caption{Zero-shot performances on experimented LLMs and datasets. Datasets above the single line are pre- LMM training data collection datasets. Confidence intervals are computed using a t-distribution. Bold text indicates significantly larger than the majority baseline using a t-test with $p=.99$. A graphical representation of this data is in Figs.~\ref{fig:experiments-zero-shot} and \ref{fig:experiments-few-shot}.}
\label{tab:zero-shot}
\end{table*}

\begin{table*}[h!]
\centering\resizebox{\textwidth}{!}{
\begin{tabular}{l||llllll||lllllll}
Dataset   & Majority                           & davinci & davinci-001 & davinci-002 & davinci-003 & GPT-3.5-T & MoE-7B & GPT-J-6B    & OPT-6.7B & BLOOM-7B      & LLama-7B & Alpaca-7B & Vicuna-7B \\ \hline \hline
        RTE &     52.7 &           50.5$\pm$3.5 &           65.0$\pm$3.2 &  \textbf{83.4$\pm$2.0} &  \textbf{85.6$\pm$1.7} &  \textbf{84.8$\pm$1.8} &           46.6$\pm$3.5 &           46.6$\pm$3.5 &           62.8$\pm$3.3 &           51.6$\pm$3.5 &           48.0$\pm$3.5 &           62.5$\pm$3.3 &  \textbf{71.8$\pm$2.9} \\
       WNLI &     56.3 &           57.7$\pm$7.0 &           46.5$\pm$7.1 &           60.6$\pm$6.8 &           71.8$\pm$5.8 &  \textbf{85.9$\pm$3.5} &           56.3$\pm$7.0 &           46.5$\pm$7.1 &           43.7$\pm$7.0 &           52.1$\pm$7.1 &           46.5$\pm$7.1 &           46.5$\pm$7.1 &           64.8$\pm$6.5 \\
       COPA &     55.0 &           47.0$\pm$5.9 &  \textbf{83.0$\pm$3.4} &  \textbf{96.0$\pm$0.9} &  \textbf{96.0$\pm$0.9} &  \textbf{97.0$\pm$0.7} &  \textbf{90.0$\pm$2.1} &           45.0$\pm$5.9 &           54.0$\pm$5.9 &           45.0$\pm$5.9 &           69.0$\pm$5.1 &           66.0$\pm$5.4 &           72.0$\pm$4.8 \\
      SST-2 &     50.9 &  \textbf{91.7$\pm$0.6} &  \textbf{92.7$\pm$0.5} &  \textbf{92.2$\pm$0.6} &  \textbf{78.2$\pm$1.3} &  \textbf{90.1$\pm$0.7} &            1.7$\pm$0.1 &  \textbf{79.5$\pm$1.3} &  \textbf{87.4$\pm$0.9} &  \textbf{84.7$\pm$1.0} &  \textbf{93.6$\pm$0.5} &  \textbf{93.2$\pm$0.5} &  \textbf{87.3$\pm$0.9} \\
       MRPC &     68.4 &           52.7$\pm$2.9 &           69.1$\pm$2.5 &           71.6$\pm$2.4 &           77.0$\pm$2.1 &           72.8$\pm$2.3 &           31.6$\pm$2.5 &  \textbf{85.3$\pm$1.5} &           67.2$\pm$2.6 &           31.6$\pm$2.5 &           69.4$\pm$2.5 &           68.4$\pm$2.5 &           53.9$\pm$2.9 \\
       QNLI &     50.5 &           51.7$\pm$0.8 &  \textbf{59.0$\pm$0.8} &  \textbf{79.0$\pm$0.5} &  \textbf{79.9$\pm$0.5} &  \textbf{84.4$\pm$0.4} &           50.6$\pm$0.8 &           49.5$\pm$0.8 &  \textbf{55.6$\pm$0.8} &           52.1$\pm$0.8 &  \textbf{57.7$\pm$0.8} &  \textbf{58.8$\pm$0.8} &  \textbf{70.3$\pm$0.7} \\
         CB &     50.0 &           50.0$\pm$8.1 &  \textbf{80.4$\pm$5.1} &           78.6$\pm$5.5 &           78.6$\pm$5.5 &  \textbf{80.4$\pm$5.1} &            0.0$\pm$0.0 &           44.6$\pm$8.1 &           41.1$\pm$7.9 &           41.1$\pm$7.9 &           71.4$\pm$6.6 &  \textbf{83.9$\pm$4.4} &           53.6$\pm$8.1 \\
        WiC &     50.0 &           51.1$\pm$2.2 &           55.6$\pm$2.2 &           57.2$\pm$2.2 &  \textbf{66.5$\pm$2.0} &  \textbf{63.2$\pm$2.1} &           50.0$\pm$2.2 &           54.9$\pm$2.2 &           50.2$\pm$2.2 &           51.3$\pm$2.2 &           50.5$\pm$2.2 &           49.8$\pm$2.2 &           52.4$\pm$2.2 \\
      BoolQ &     62.2 &           55.8$\pm$0.9 &  \textbf{79.5$\pm$0.6} &  \textbf{87.1$\pm$0.4} &  \textbf{88.4$\pm$0.4} &  \textbf{85.1$\pm$0.5} &           37.9$\pm$0.9 &           62.9$\pm$0.9 &  \textbf{66.9$\pm$0.8} &           52.6$\pm$1.0 &  \textbf{77.8$\pm$0.7} &  \textbf{73.2$\pm$0.7} &  \textbf{76.0$\pm$0.7} \\\cline{1-1} \cline{2-4} \cline{8-9}
 StrategyQA &     53.3 &           52.4$\pm$3.9 &           58.5$\pm$3.8 &           \multicolumn{1}{|l}{62.4$\pm$3.6} &  \textbf{70.3$\pm$3.2} &  \textbf{69.0$\pm$3.3} &           48.5$\pm$3.9 &           45.0$\pm$3.8 &           \multicolumn{1}{|l}{52.8$\pm$3.9} &           49.8$\pm$3.9 &           53.3$\pm$3.9 &           61.1$\pm$3.7 &           56.8$\pm$3.8 \\
    MTSC-MT &     50.7 &           40.0$\pm$1.5 &           43.2$\pm$1.5 &  \multicolumn{1}{|l}{\textbf{61.0$\pm$1.4}} &  \textbf{68.4$\pm$1.3} &  \textbf{70.7$\pm$1.3} &            0.1$\pm$0.0 &           36.7$\pm$1.4 &           \multicolumn{1}{|l}{24.1$\pm$1.1} &            2.9$\pm$0.2 &           48.3$\pm$1.5 &  \textbf{59.2$\pm$1.5} &           54.3$\pm$1.5 \\
    MTSC-RW &     39.7 &           33.2$\pm$1.5 &  \textbf{52.9$\pm$1.7} &  \multicolumn{1}{|l}{\textbf{66.8$\pm$1.5}} &  \textbf{64.6$\pm$1.6} &  \textbf{69.4$\pm$1.5} &            0.1$\pm$0.0 &           31.0$\pm$1.5 &           \multicolumn{1}{|l}{30.8$\pm$1.5} &            3.1$\pm$0.2 &           41.4$\pm$1.7 &  \textbf{55.2$\pm$1.7} &  \textbf{55.7$\pm$1.7} \\\cline{5-7} \cline{10-10}
  NLI4Wills &     55.7 &           47.1$\pm$3.7 &           30.2$\pm$3.1 &            5.1$\pm$0.7 &           28.2$\pm$3.0 &           36.5$\pm$3.4 &            0.4$\pm$0.1 &           21.6$\pm$2.5 &           24.3$\pm$2.7 &           \multicolumn{1}{|l}{54.9$\pm$3.6} &           56.9$\pm$3.6 &           17.6$\pm$2.1 &           19.2$\pm$2.3 \\ \cline{11-14}
        CREPE &     72.8 &           60.9$\pm$1.2 &           44.9$\pm$1.3 &           73.8$\pm$1.0 &           70.9$\pm$1.1 &           62.2$\pm$1.2 &           67.7$\pm$1.1 &           72.8$\pm$1.0 &           72.8$\pm$1.0 &           14.8$\pm$0.7 &           71.2$\pm$1.1 &           72.8$\pm$1.0 &           72.8$\pm$1.0 \\
       FOMC &     49.4 &           40.7$\pm$2.5 &           54.4$\pm$2.6 &           55.2$\pm$2.6 &  \textbf{61.7$\pm$2.5} &  \textbf{63.5$\pm$2.4} &           25.0$\pm$2.0 &           49.4$\pm$2.6 &           49.4$\pm$2.6 &           42.3$\pm$2.6 &           50.2$\pm$2.6 &           52.8$\pm$2.6 &           50.0$\pm$2.6 \\
    NewsMet &     52.3 &           48.0$\pm$2.5 &           51.3$\pm$2.5 &           49.5$\pm$2.5 &           50.2$\pm$2.5 &           56.0$\pm$2.4 &           39.4$\pm$2.4 &           47.7$\pm$2.5 &           52.5$\pm$2.5 &           47.7$\pm$2.5 &           52.3$\pm$2.5 &           50.9$\pm$2.5 &           52.0$\pm$2.5 \\
\end{tabular}}
\caption{Few-shot performances on GPT-series models. Datasets above the single line are pre- LMM training data collection datasets. Confidence intervals are computed using a t-distribution. Bold text indicates significantly larger than the majority baseline using a t-test with $p=.99$. A graphical representation of this data is in Figs.~\ref{fig:experiments-zero-shot} and \ref{fig:experiments-few-shot}.}
\label{tab:few-shot}
\end{table*}

\section{Additional Figures}

\begin{figure*}[h!]
        \centering
        \begin{subfigure}[b]{0.475\textwidth}
            \centering
            \includegraphics[scale=0.45]{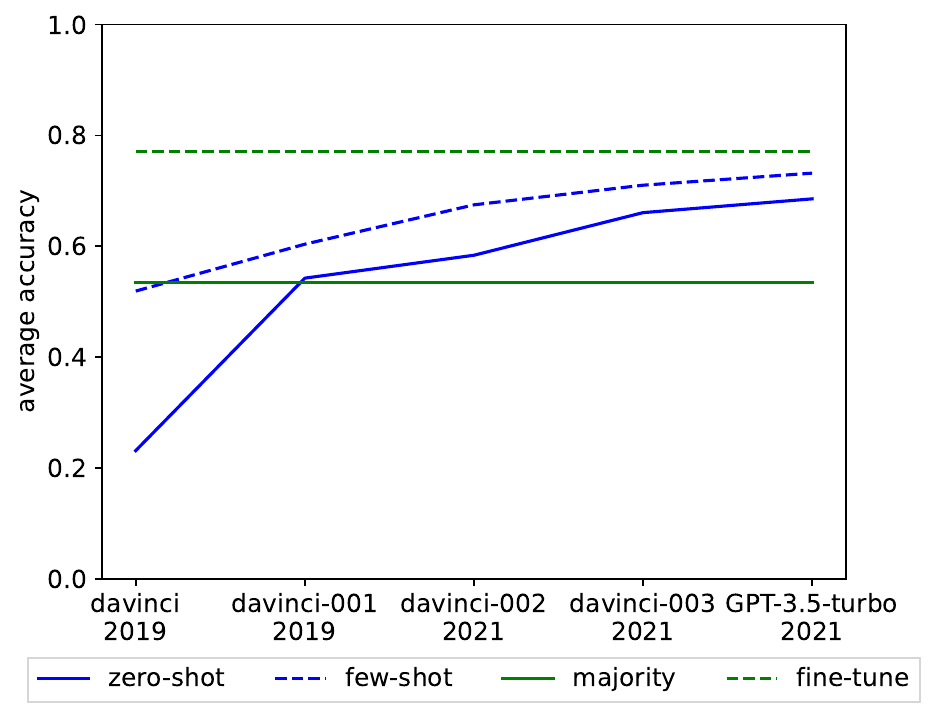}
            \caption[]%
            {{GPT-3 series}}    
            \label{fig:GPT-datasets}
        \end{subfigure}
        \begin{subfigure}[b]{0.475\textwidth}   
            \centering 
            \includegraphics[scale=0.45]{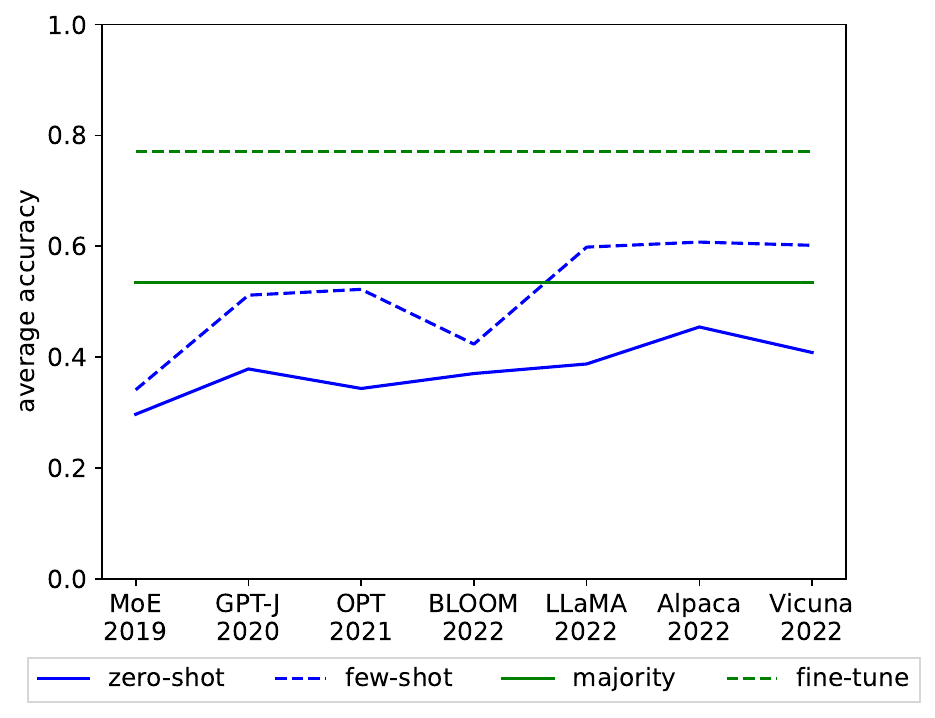}
            \caption[]%
            {{Open LLMs}}    
            \label{fig:Open-datasets}
        \end{subfigure}
        \caption[]
        {Average performance across all datasets for GPT-3 series and open LLMs. In the $x$ axis, LLMs are ordered chronologically by training data collection date, and the collection year is listed below the LLM.} 
        \label{fig:experiments-average-all}
    \end{figure*}


\begin{figure*}[t]
        \centering
        \begin{subfigure}[b]{0.45\textwidth}
            \centering
            \includegraphics[scale=0.38]{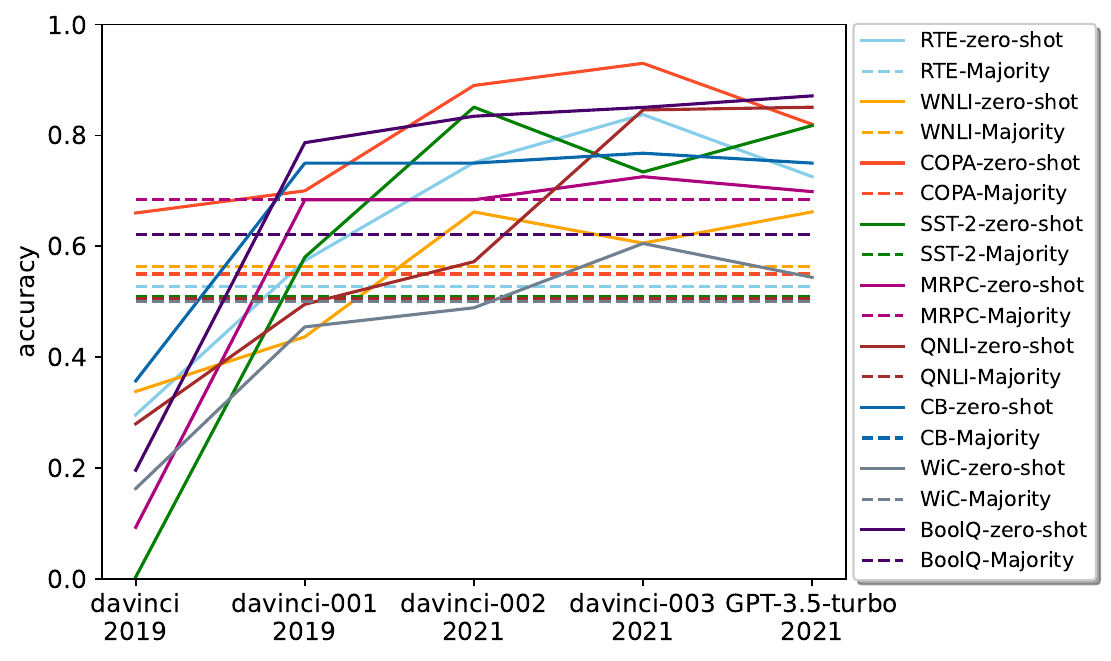}
            \caption[]%
            {{GPT zero-shot performance on pre-2021 datasets.\jmfb{patterns in text-davinci-003 are mirrored in Vicuna (mostly)}}}    
            \label{fig:Zero shot performance for old datasets}
        \end{subfigure}
        \hfill
        \begin{subfigure}[b]{0.45\textwidth}   
            \centering 
            \includegraphics[scale=0.38]{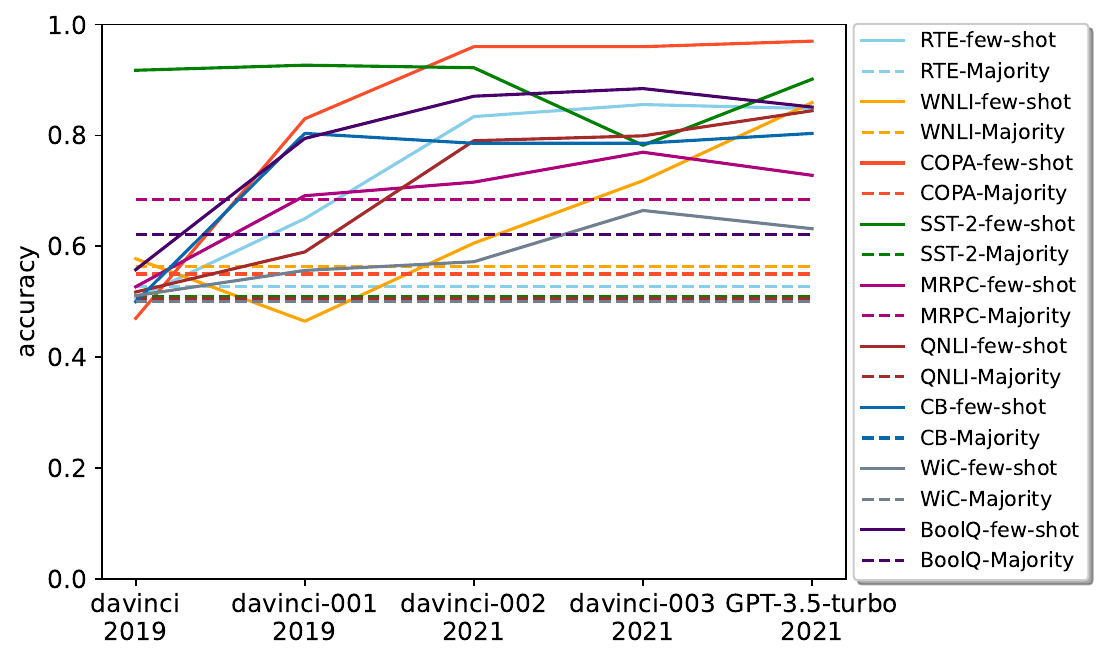}
            \caption[]%
            {{GPT few-shot performance on pre-2021.}}    
            \label{fig:Few shot performance for old datasets}
        \end{subfigure}
        \vskip\baselineskip
        \begin{subfigure}[b]{0.45\textwidth}
            \centering
            \includegraphics[scale=0.38]{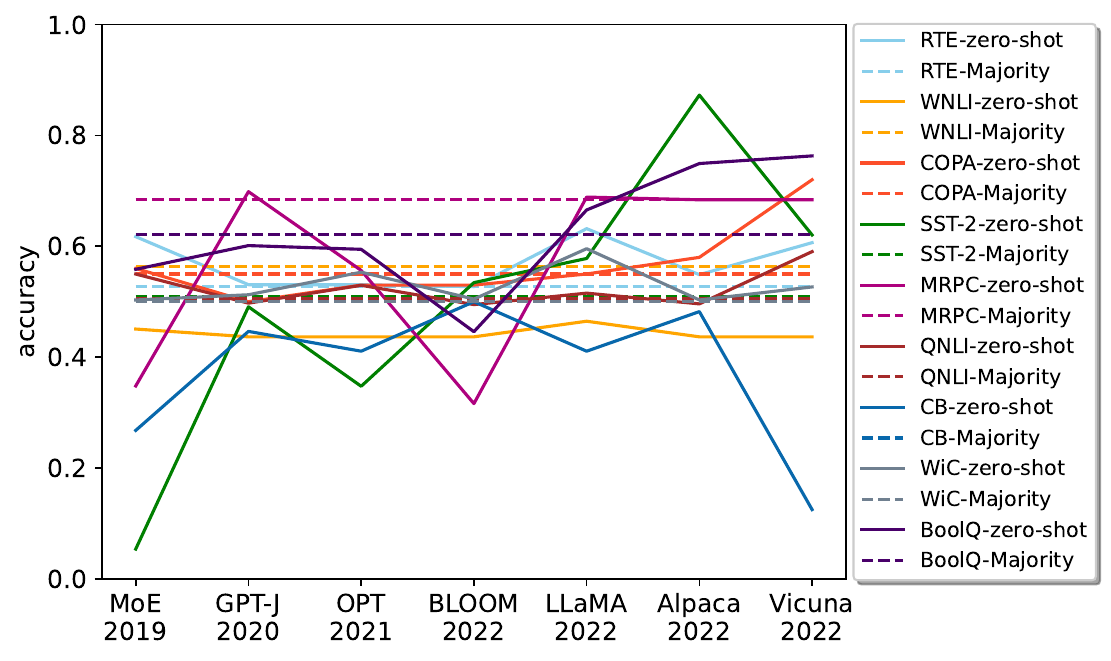}
            \caption[]%
            {{Open LLM zero-shot performance on pre-2021.\jmfb{My guess is OPT probably does better than davinci because of the weighting of the datasets during training}}}    
            \label{fig:Zero shot performance for old datasets}
        \end{subfigure}
        \hfill
        \begin{subfigure}[b]{0.45\textwidth}   
            \centering 
            \includegraphics[scale=0.38]{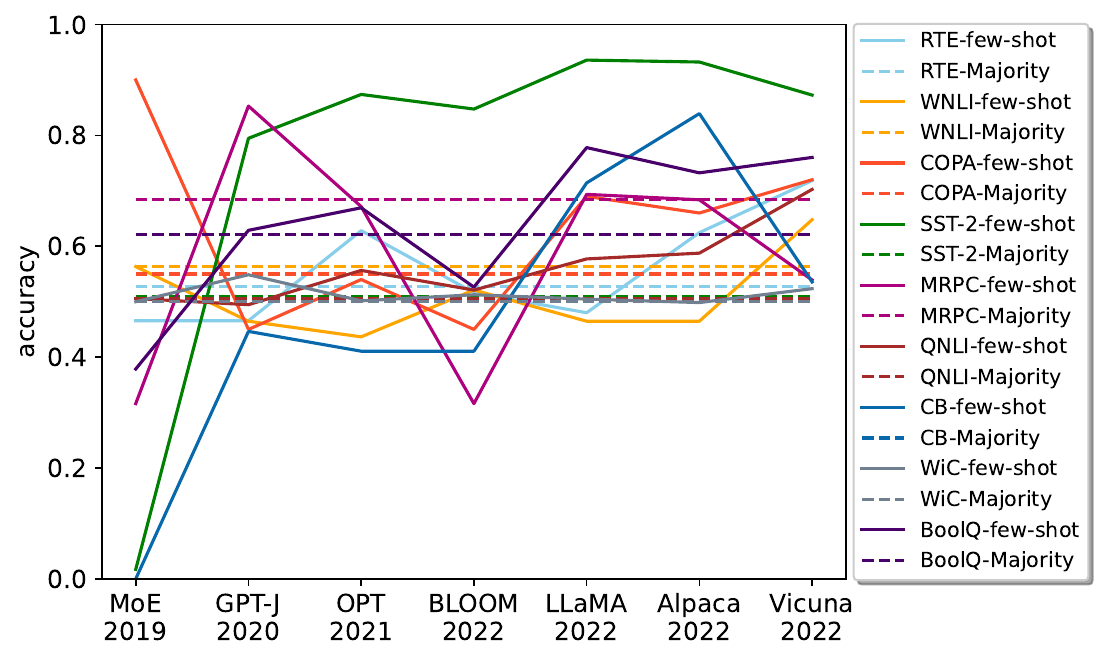}
            \caption[]%
            {Open LLM few-shot performance on pre-2021. \jmf{five shots}\jmfb{why does is it so good on sst? I think because it's true few shot learning }}    
            \label{fig:Few shot performance for old datasets}
        \end{subfigure}
        \caption[]
        {Performance on pre-2021 datasets. In the $x$ axis, LLMs are ordered chronologically. Dotted lines are majority baselines.} 
        \label{fig:experiments-zero-shot}
    \end{figure*}

\begin{figure*}[b]
        \centering
        \begin{subfigure}[b]{0.45\textwidth}  
            \centering 
            \includegraphics[scale=0.38]{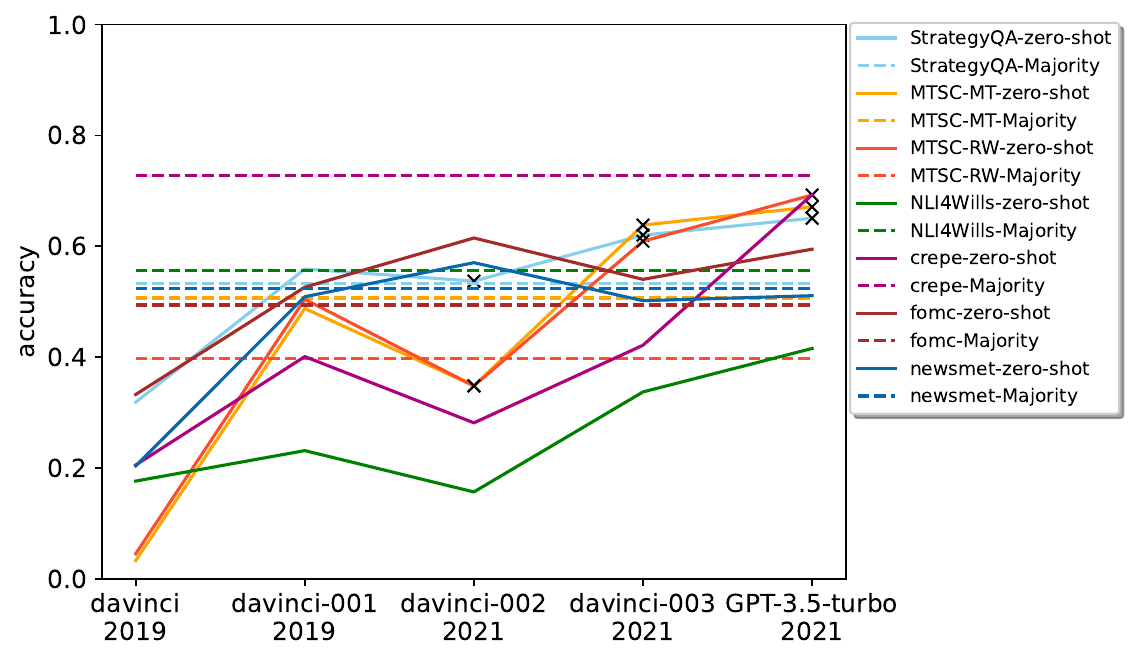}
            \caption[]%
            {GPT zero-shot performance on post-2021 datasets.}    
            \label{fig:Zero shot performance for new datasets}
        \end{subfigure}
        \hfill
        \begin{subfigure}[b]{0.45\textwidth}   
            \centering 
            \includegraphics[scale=0.38]{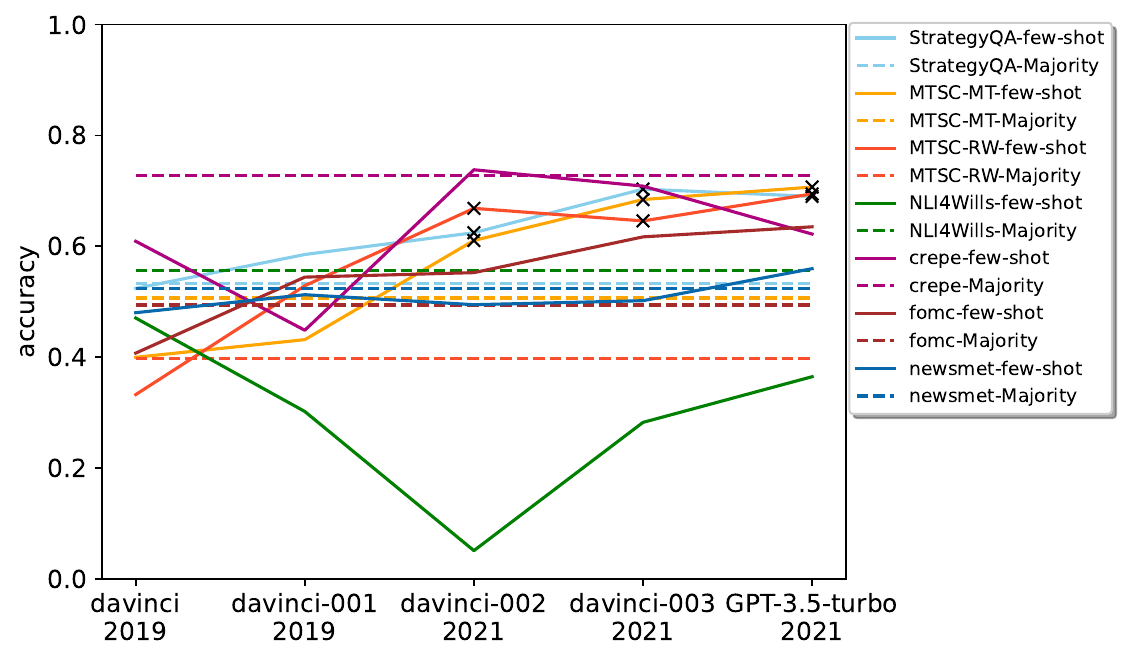}
            \caption[]%
            {GPT few-shot performance on post-2021 datasets.\jmfb{say in the text, that davinci-002 and up could have seen some of these datasets}}    
            \label{fig:Few shot performance for new datasets}
        \end{subfigure}
        \vskip\baselineskip
        \begin{subfigure}[b]{0.45\textwidth}  
            \centering 
            \includegraphics[scale=0.38]{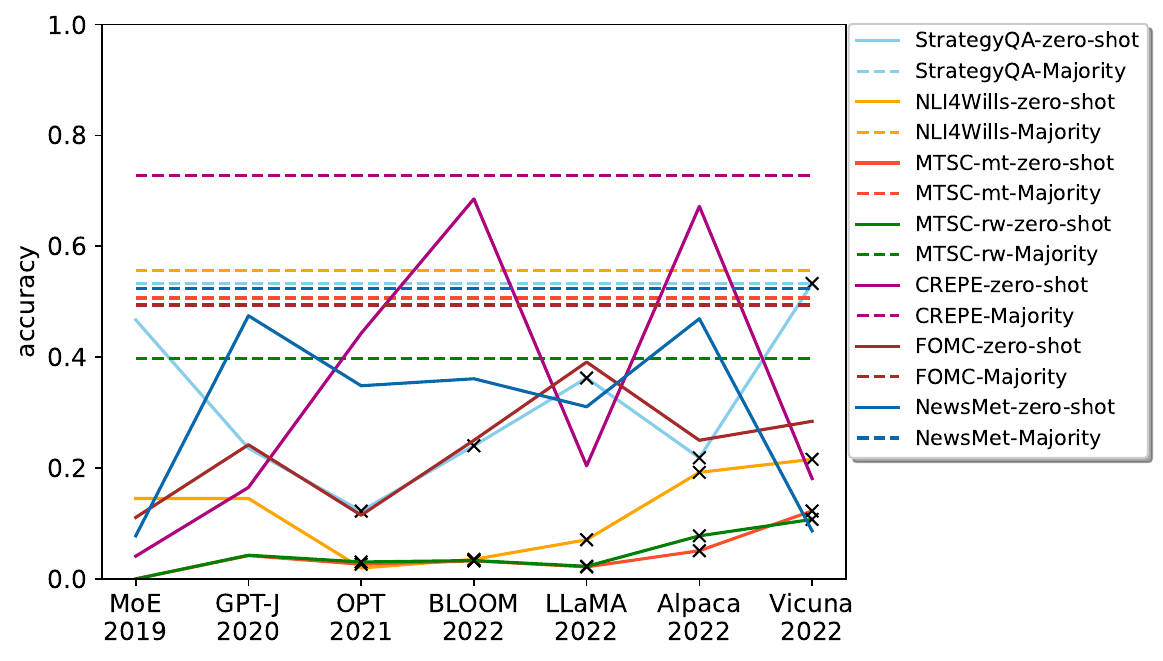}
            \caption[]%
            {Open LLM zero-shot performance on post-2021.}    
            \label{fig:Zero shot performance for new datasets}
        \end{subfigure}
        \hfill
        \begin{subfigure}[b]{0.45\textwidth}   
            \centering 
            \includegraphics[scale=0.38]{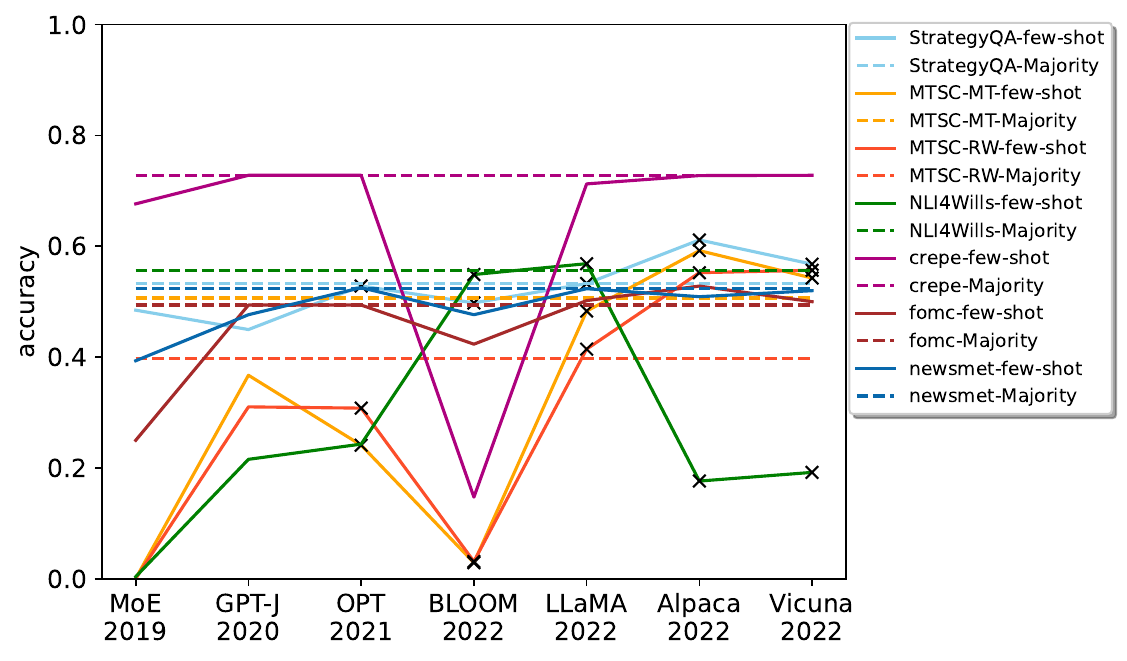}
            \caption[]%
            {Open LLM few-shot performance on post-2021.}    
            \label{fig:Few shot performance for new datasets}
        \end{subfigure}
        \caption[]
        {Performance on post-2021 datasets. In the $x$ axis, LLMs are ordered chronologically. Dotted lines are majority baselines. "x" indicates the model may have seen the dataset based on the date: the model training data collection date is after the dataset release date.} 
        \label{fig:experiments-few-shot}
    \end{figure*}

\clearpage

\onecolumn
\section{Prompt Examples for Each Task}
\label{app:prompt_examples}

In this section we give examples of zero-shot prompts for each task.

\begin{table*}[h!]
\begin{tabular}{l}
Task: MRPC \\ \hline
Prompting Inputs: \\ \hline
\begin{tabular}[c]{@{}l@{}}He said the foodservice pie business doesn 't fit the company 's long-term growth strategy\\ .
" The foodservice pie business does not fit our long-term growth strategy .
Are the previous\\ two sentences are paraphrased, respond as yes or no?\\ \end{tabular} \\ \hline
Expected Outputs: \\ \hline
Yes
\end{tabular}
\end{table*}

\begin{table*}[h!]
\begin{tabular}{l}
Task: BOOLQ \\ \hline
Prompting Inputs: \\ \hline
\begin{tabular}[c]{@{}l@{}}Ethanol fuel -- All biomass goes through at least some of these steps: it needs\\ to be grown, collected, dried, fermented, distilled, and burned. All of these steps require resources\\ and an infrastructure. The total amount of energy input into the process compared to the\\ energy released by burning the resulting ethanol fuel is known as the energy balance (or\\ ``energy returned on energy invested''). Figures compiled in a 2007 report \\  by National Geographic Magazine point to modest results for corn ethanol produced in the US: \\ one unit of fossil-fuel energy is required to create 1.3 energy units from the resulting ethanol. \\ The energy balance for sugarcane ethanol produced in Brazil is more favorable, \\ with one unit of fossil-fuel energy required to create 8 from the ethanol. \\ Energy balance estimates are not easily produced, thus\\ numerous such reports have been generated that are contradictory. \\ For instance, a separate survey reports that production of ethanol from sugarcane, \\ which requires a tropical climate to grow productively, \\ returns from 8 to 9 units of energy for each unit expended, as compared to corn,\\which only returns about 1.34 units of fuel energy for each unit of energy expended.\\ A 2006 University of California Berkeley study, after analyzing six separate studies, \\ concluded that producing ethanol from corn uses much less petroleum than producing gasoline. \\
Does ethanol take more energy make that produces, respond as yes or no?\\ \end{tabular} \\ \hline
Expected Outputs: \\ \hline
No
\end{tabular}
\end{table*}

\begin{table*}[h!]
\begin{tabular}{l}
Task: SST \\ \hline
Prompting Inputs: \\ \hline
\begin{tabular}[c]{@{}l@{}}Classify the sentiment:it 's a charming and often affecting journey . \\ \end{tabular} \\ \hline
Expected Outputs: \\ \hline
Positive
\end{tabular}
\end{table*}

\begin{table*}[h!]
\begin{tabular}{l}
Task: QQP \\ \hline
Prompting Inputs: \\ \hline
\begin{tabular}[c]{@{}l@{}}Why are African-Americans so beautiful?
Why are hispanics so beautiful? \\ 
Are the previous two sentences are paraphrased, respond as yes or no?\\ \end{tabular} \\ \hline
Expected Outputs: \\ \hline
No
\end{tabular}
\end{table*}

\begin{table*}[h!]
\begin{tabular}{l}
Task: QNLI \\ \hline
Prompting Inputs: \\ \hline
\begin{tabular}[c]{@{}l@{}}Entailment: if the context contains the answer to the question, then it is entailment. \\
Question: What came into force after the new constitution was herald? \\
Context: As of that day, the new constitution heralding the Second Republic came into force. \\
Is the context entailment, Yes or No?\\ \end{tabular} \\ \hline
Expected Outputs: \\ \hline
Yes
\end{tabular}
\end{table*}

\begin{table*}[h!]
\begin{tabular}{l}
Task: WNLI \\ \hline
Prompting Inputs: \\ \hline
\begin{tabular}[c]{@{}l@{}}Entailment: if the premise is true, then the hypothesis must be true.
Premise: The drain is\\ clogged with hair. It has to be cleaned.
Hypothesis: The hair has to be cleaned.
Is the\\ hypothesis entailment?\\ \end{tabular} \\ \hline
Expected Outputs: \\ \hline
No
\end{tabular}
\end{table*}

\begin{table*}[h!]
\begin{tabular}{l}
Task: RTE \\ \hline
Prompting Inputs: \\ \hline
\begin{tabular}[c]{@{}l@{}}Entailment: if the premise is true, then the hypothesis must be true.
Premise: Dana Reeve, the\\ widow of the actor Christopher Reeve, has died of lung cancer at age 44, according\\ to the Christopher Reeve Foundation.
Hypothesis: Christopher Reeve had an accident. \\
Is the hypothesis entailment? \end{tabular} \\ \hline
Expected Outputs: \\ \hline
No
\end{tabular}
\end{table*}

\begin{table*}[h!]
\begin{tabular}{l}
Task: CB \\ \hline
Prompting Inputs: \\ \hline
\begin{tabular}[c]{@{}l@{}}Please identify whether the premise entails the hypothesis. \\ The answer should be exact ’yes’, ’no’ or ’neutral’.
\\
premise: Valence the void-brain, Valence the virtuous valet. \\ Why couldn't the figger choose his own portion of titanic anatomy to shaft? \\ Did he think he was helping?
\\ hypothesis: Valence was helping
\\ answer:\\ \end{tabular} \\ \hline
Expected Outputs: \\ \hline
No
\end{tabular}
\end{table*}

\begin{table*}[h!]
\begin{tabular}{l}
Task: COPA \\ \hline
Prompting Inputs: \\ \hline
\begin{tabular}[c]{@{}l@{}}The man turned on the faucet. What happened as a result?
1. The toilet filled with\\ water.
2. Water flowed from the spout.
Which one, 1 or 2?\\ \end{tabular} \\ \hline
Expected Outputs: \\ \hline
2
\end{tabular}
\end{table*}

\begin{table*}[h!]
\begin{tabular}{l}
Task: WIC \\ \hline
Prompting Inputs: \\ \hline
\begin{tabular}[c]{@{}l@{}}An emerging professional class.
Apologizing for losing your temper, \\ even though you were badly provoked, showed real class. \\
Does the word class have the same word sense, Yes or No?\\ \end{tabular} \\ \hline
Expected Outputs: \\ \hline
No
\end{tabular}
\end{table*}

\begin{table*}[h!]
\begin{tabular}{l}
Task: STRATEGYQA \\ \hline
Prompting Inputs: \\ \hline
\begin{tabular}[c]{@{}l@{}}Q: Will the Albany in Georgia reach a hundred thousand occupants before the one in\\ New York?
\\ A: The answer (Yes or No) is\\ \end{tabular} \\ \hline
Expected Outputs: \\ \hline
No
\end{tabular}
\end{table*}

\begin{table*}[t!]
\begin{tabular}{l}
Task: NLI4WILLS \\ \hline
Prompting Inputs: \\ \hline
\begin{tabular}[c]{@{}l@{}}Law: 32-3-111. Specifically devised or bequeathed property. (a) A specific legatee or devisee has a\\ right to the specifically gifted or devised property in the testator's estate at death or\\ if the property has been disposed of and a contrary intention is not manifest during\\ the testator's lifetime: (1) Any balance of the purchase price, together with any security interest,\\ owing from a purchaser to the testator at death by reason of sale of the\\ property; (2) Any amount of a condemnation award for the taking of the property unpaid\\ at death; (3) Any proceeds unpaid at death on fire or casualty insurance on, or\\ other recovery for injury to, the property; and (4) Property owned by the testator at\\ death and acquired as a result of foreclosure, or obtained in lieu of foreclosure, of\\ the security interest for a specifically devised obligation. \\
Condition: The testator and his wife didn't divorce until the testator's death, \\ and the testator's wife survived the testator. \\
Statement: I give, devise and bequeath all my property, real, personal and mixed, \\ of whatever kind and nature and wheresoever situated, to my wife, [Person-2], \\ if she survives me. \\
Given the law and condition, check the statement for validity (output Support, Refute, or Unrelated). \\
Answer:\\ \end{tabular} \\ \hline
Expected Outputs: \\ \hline
Refute
\end{tabular}
\end{table*}

\begin{table*}[t!]
\begin{tabular}{l}
Task: NEWSMTSC-RW \\ \hline
Prompting Inputs: \\ \hline
\begin{tabular}[c]{@{}l@{}}Classify the sentiment of the sentence concerning target Mr. Trump as positive, neutral, or negative:\\ A group of congressional Democrats said Wednesday that they will ask Congress to take the\\ rare step of officially censuring Mr. Trump.\\ \end{tabular} \\ \hline
Expected Outputs: \\ \hline
negative
\end{tabular}
\end{table*}

\begin{table*}[h!]
\begin{tabular}{l}
Task: NEWSMTSC-MT \\ \hline
Prompting Inputs: \\ \hline
\begin{tabular}[c]{@{}l@{}}Classify the sentiment of the sentence concerning target Hillary Clinton’s as positive, neutral, or negative:\\ While White House officials said in the days after Comey's dismissal that it was largely\\ the result of a memo written by Deputy Attorney General Rod J. Rosenstein criticizing the\\ FBI director's handling of the investigation into Hillary Clinton’s use of a private email server\\ when she was secretary of state, Trump suggested in the NBC interview that the Russian\\ investigation played a role in his decision.\\ \end{tabular} \\ \hline
Expected Outputs: \\ \hline
negative
\end{tabular}
\end{table*}

\begin{table*}[h!]
\begin{tabular}{l}
Task: Spider without schema\\ \hline
Prompting Inputs: \\ \hline
\begin{tabular}[c]{@{}l@{}}Create a SQL request to how many singers do we have?\\
 SELECT\\ \end{tabular} \\ \hline
Expected Outputs: \\ \hline
SELECT count(*) FROM singer
\end{tabular}
\end{table*}

\begin{table*}[t!]
\begin{tabular}{l}
Task: Spider with schema \\ \hline
Prompting Inputs: \\ \hline
\begin{tabular}[c]{@{}l@{}}\#\#\# Postgres SQL tables, with their properties: \\
\# \\
\# stadium(Stadium\_ID, Location, Name, Capacity, Highest, Lowest, Average) \\
\# singer(Singer\_ID, Name, Country, Song\_Name, Song\_release\_year, Age, Is\_male) \\
\# concert(concert\_ID, concert\_Name, Theme, Stadium\_ID, Year) \\
\# singer\_in\_concert(concert\_ID, Singer\_ID) \\
\# \\
\#\#\# A query to how many singers do we have? \\
SELECT\\ \end{tabular} \\ \hline
Expected Outputs: \\ \hline
SELECT count(*) FROM singer
\end{tabular}
\end{table*}

\begin{table*}[t!]
\begin{tabular}{l}
Task: FOMC \\ \hline
Prompting Inputs: \\ \hline
\begin{tabular}[c]{@{}l@{}}
Classify the following sentence from FOMC into 'HAWKISH', 'DOVISH', or 'NEUTRAL' class. \\
Label 'HAWKISH' if it is corresponding to tightening of the monetary 
policy, \\ 'DOVISH' if it is corresponding to easing of the monetary policy,
, or 'NEUTRAL' if the stance is neutral. \\
The sentence: During the past several years, workers across the wage distribution--not just at the \\ upper end--have seen noticeable increases in the inflation-adjusted value of their wages.
Label: \\
\end{tabular} \\ \hline
Expected Outputs: \\ \hline
Hawkish
\end{tabular}
\end{table*}

\begin{table*}[t!]
\begin{tabular}{l}
Task: CREPE \\ \hline
Prompting Inputs: \\ \hline
\begin{tabular}[c]{@{}l@{}}
Question: Why does a cold cause your voice to get deeper?\\
Comment:  Swelling of the vocal folds makes them heavier and that causes them to vibrate at lower (deeper) frequencies. \\ If you look at a guitar or any string instrument you will notice the thicker strings are the lower notes.\\
Does comment have false presuppositions to the question, Yes or No? \\
\end{tabular} \\ \hline
Expected Outputs: \\ \hline
No
\end{tabular}
\end{table*}

\begin{table*}[t!]
\begin{tabular}{l}
Task: NewsMet \\ \hline
Prompting Inputs: \\ \hline
\begin{tabular}[c]{@{}l@{}}
Classify the following sentence into 'literal', or 'metaphorical' class. 
Label 'literal' if it is not metaphorical. \\ Label 'metaphorical' if it is metaphorical. \\
The sentence: President Donald Trump kicks CNN reporter out of Oval Office\\
Label: \\
\end{tabular} \\ \hline
Expected Outputs: \\ \hline
metaphorical
\end{tabular}
\end{table*}

\newpage
\clearpage

\section{Prompts for Task Example Extraction}
\label{app:prompt_extraction}

\begin{table}[h!]
\begin{tabular}{l|p{5.5in}}
Task &
  Prompt used \\ \hline \hline
RTE & 
  Generate several training examples for Recognizing Textual Entailment dataset including premise and hypothesis with entailment and not\_entailment as labels. \\ \hline
WNLI & 
  Generate several training examples for Winograd Schema Natural Language Inference dataset including premise and hypothesis with entailment and not\_entailment as labels.   \\ \hline
COPA & 
  Generate several training examples for Choice of Plausible Alternatives (COPA) dataset including premise and choices as input with 0 or 1 as labels. \\ \hline
SST-2 &
  Generate several training examples for sentiment analysis task with positve and negative as labels \\ \hline
MRPC &
  Generate several training examples for Microsoft Research Paraphrase Corpus task. \\ \hline
QNLI &  
  Generate several training examples for Question answering  Natural Language Inference dataset using question answer pairs with entailment and not\_entailment as labels.   \\ \hline
CB &  
  Generate several training examples for CommitmentBank Natural Language Inference dataset including premise and hypothesis as input with entailment, neutral, as contradiction labels.  \\ \hline
WiC &  
  Generate several training examples for The Word-in-Context (WiC) Dataset task including 2 sentences and a word in both sentences as input with true or false as labels.   \\ \hline
BoolQ &  
  Generate several training examples for BoolQ dataset which is a question answering dataset for yes/no questions including passage and question as input with yes or no as labels.  \\ \hline \hline
StrategyQA &  
  Generate several training examples for StrategyQA task which is a question-answering task focusing on open-domain questions where the required reasoning steps are implicit in the question and should be inferred using a strategy.  Generate with a question and reasoning steps as input and Yes or No as Labels.  \\ \hline
NewsMTSC &  
  Generate several training examples for Multi-Target-dependent Sentiment Classification in Political News Articles including a sentence and a target in the sentence as input with positive and negative as labels.  \\ \hline
NLI4Wills &  
  Generate several training examples for the validity evaluation of the legal will statements including statement,  conditions and law as input with support, refute, or unrelated as labels.  \\ \hline
CREPE &  
  Generate several training examples for a QA task containing a natural distribution of presupposition failures for questions with whether there is any false presuppositions including question and comment as input with true or false as labels  \\ \hline
FOMC &  
  Generate several training examples for Federal Open Market Committee (FOMC) dataset for a measure of monetary policy stance task including sentence from FOMC document as input with Dovish, Hawkish or Neutral as labels.  \\ \hline
NewsMet &  
  Generate several training examples from NewsMet, a large high-quality contemporary dataset of news headlines hand-annotated with metaphorical verbs with a task to detect if the headline is metaphorical including a headline  sentence as input with 0 or 1 as labels to represent metaphorical or not metaphorical.  \\ \hline
\end{tabular}
\caption{Prompts used for each task for task example extraction.}
\label{tab:task-extraction-prompts}
\end{table}

\end{document}